\documentclass[10pt,twocolumn,letterpaper]{article}

\usepackage{cvpr}
\usepackage{times}
\usepackage{epsfig}
\usepackage{graphicx}
\usepackage{amsmath}
\usepackage{amssymb}

\usepackage{multirow}
\usepackage{color}
\usepackage{subfigure}

\usepackage{algorithm}
\usepackage{algorithmic}
\usepackage[pagebackref=true,breaklinks=true,letterpaper=true,colorlinks,bookmarks=false]{hyperref}

\cvprfinalcopy % *** Uncomment this line for the final submission

\def\cvprPaperID{0406} % *** Enter the CVPR Paper ID here

\ifcvprfinal\fi
\begin{document}

%%%%%%%%% TITLE
\title{Location-Aware Feature Selection Text Detection Network}

\author{Zengyuan Guo$^{1}$\thanks{Equal contribution}\hspace{4mm}Zilin Wang$^{1*}$
	\hspace{4mm}Zihui Wang$^{1}$\thanks{Corresponding author:\textit{ }zhwang@dlut.edu.cn}
	\hspace{4mm}Wanli Ouyang$^{2}$\hspace{4mm}Haojie Li$^{1}$\hspace{4mm}Wen Gao$^{3}$\\
	$^{1}$Dalian University of Technology \hspace{5mm}$^{2}$The University of Sydney\hspace{5mm}$^{3}$Peking University}
% \and

\maketitle
\begin{abstract}
Regression–based text detection methods have already achieved promising performances with simple network structure and high efficiency. However, they are behind in accuracy comparing with recent segmentation-based text detectors. In this work, we discover that one important reason to this case is that regression-based methods usually utilize a fixed feature selection way, i.e. selecting features in a single location or in neighbor regions, to predict components of the bounding box, such as the distances to the boundaries or the rotation angle. The features selected through this way sometimes are not the best choices for predicting every component of a text bounding box and thus degrade the accuracy performance. To address this issue, we propose a novel Location Aware feature Selection text detection Network (LASNet). LASNet selects suitable features from different locations to separately predict the five components of a bounding box and gets the final bounding box through the combination of these components. Specifically, instead of using the classification score map to select one feature for predicting the whole bounding box as most of the existing methods did, the proposed LASNet first learns five new confidence score maps to indicate the prediction accuracy of the five bounding box components, respectively. Then, a Location-Aware Feature Selection mechanism (LAFS) is designed to weightily fuse the top-$K$ prediction results for each component according to their confidence score, and to combine the all five fused components into a final bounding box. As a result, LASNet predicts the more accurate bounding boxes by using a learnable feature selection way. The experimental results demonstrate that our LASNet achieves state-of-the-art performance with single-model and single-scale testing, outperforming all existing regression-based detectors.
\end{abstract}




\maketitle

\begin{figure}[tbp]
	\begin{center}
		\centering
		\subfigure[Without LAFS]{
			\begin{minipage}[t]{0.5\linewidth}
				\centering
				\includegraphics[width=1.7in]{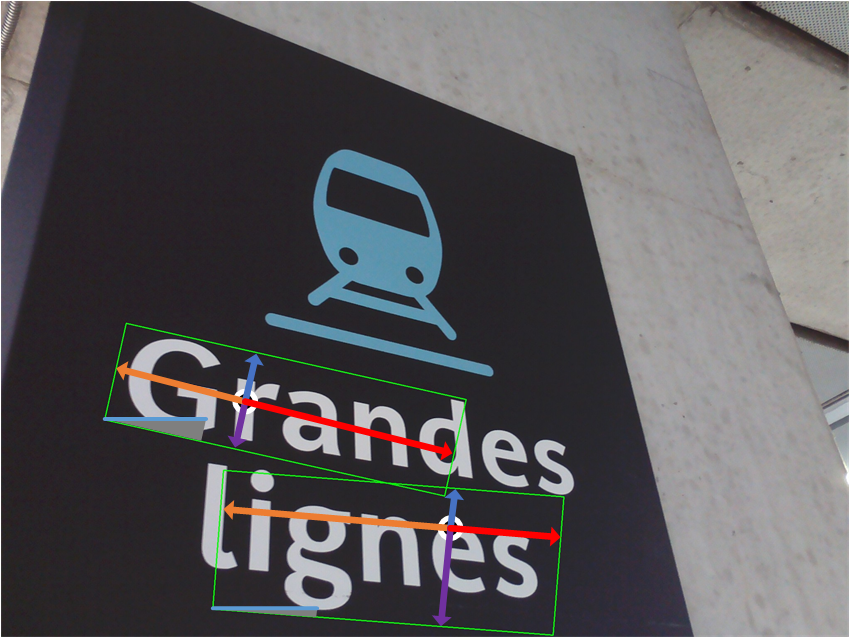}
				%\caption{Bounding box from EAST}
			\end{minipage}%
		}%
		\subfigure[With LAFS]{
			\begin{minipage}[t]{0.5\linewidth}
				\centering
				\includegraphics[width=1.7in]{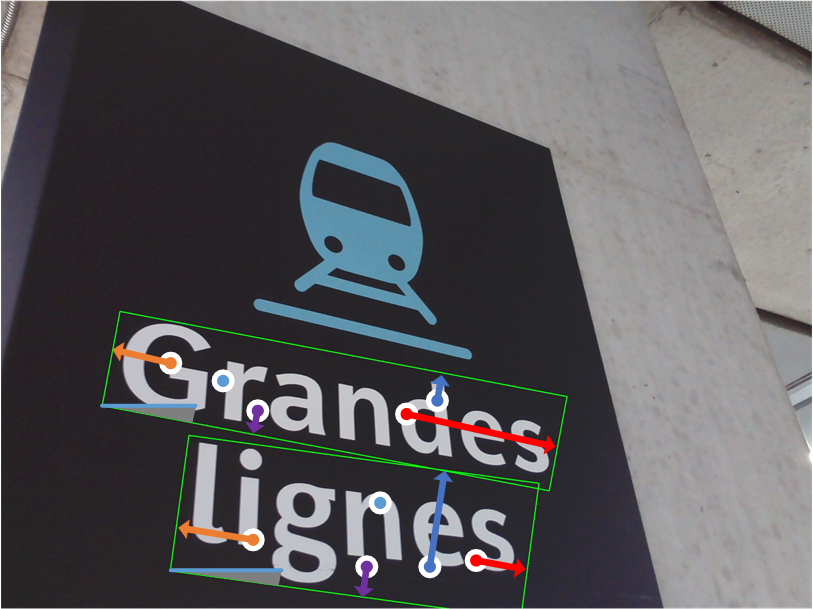}
				%\caption{Bounding box from LASNet}
			\end{minipage}%
		}%
		\centering
		\caption{(a) The bounding boxes predicted by features from a single location is inaccurate. (b) Unbinding the features used to predict different components via LAFS obviously improves the accuracy of the bounding box. The color line refers to the distance or angle predicted by the features of the corresponding circle points.}
		\label{figure1}
	\end{center}
\end{figure}

\section{Introduction}
Scene text detection based on deep learning has drawn much attention because it serves as a prerequisite for various downstream applications like self-driving car, instance translation, scene understanding and product search. Nowadays, segmentation-based text detection methods \cite{deng2018pixellink,long2018textsnake,tian2019learning,li2018shape,baek2019character} have achieved competitive performance. However, they require a complex post-process to determine the region of each text instance precisely. In contrast, direct regression-based approaches \cite{EAST,zhong2019anchor,liu2018fots,liao2018rotation,yang2018inceptext} perform boundary regression by predicting the offsets from a given point, with the advantages of simple network structure, high efficiency and no need of complex post-process. Unfortunately, it’s hard for them to accurately locate the bounding boxes, especially for large and long text.

In this work, we argue that one important reason of the inaccurate bounding boxes regression in regression-based approaches is that the features they used to predict different components of the bounding box, such as distance to the boundary or rotation angle, are always fixed from a single location or the locations close to the target components, while they neglect the fact that features from different locations of feature map are more suitable to predict different components of bounding box. For example, \cite{EAST,li2018pixel,zhong2019anchor} use features from a single location to predict all components. The prediction of different components of a bounding box was bounded to the same location. As shown in Fig. \ref{figure1}, the features from a fixed location, i.e. the circle point within the text instance, is selected to predict all components of the bounding box. In this case, the right boundary of the upper text instance and the left boundary of the under one are inaccurately predicted. Fig. \ref{figure1} (b) shows the results of our proposed LAFS. Through unbinding the features for components prediction and selecting five suitable features (i.e. five circle points in Fig. \ref{figure1} (b)) for five components, respectively, the predicted bounding box seems obviously more accurate than the results in Fig. \ref{figure1} (a).

\begin{figure}[htbp]
	\centering
	\subfigure[text instance]{
		\begin{minipage}[t]{0.3\linewidth}
			\centering
			\includegraphics[width=1in, height=1in]{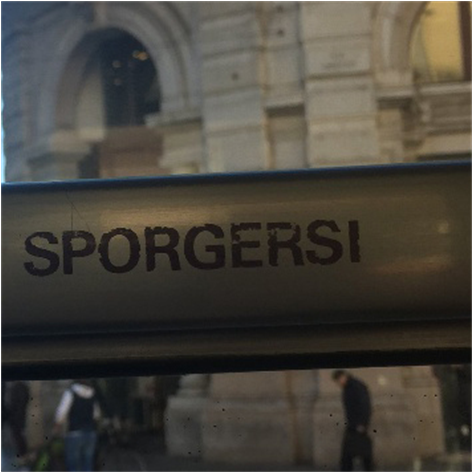}
			%\caption{fig1}
		\end{minipage}%
	}%
	\subfigure[top channel]{
		\begin{minipage}[t]{0.3\linewidth}
			\centering
			\includegraphics[width=1in, height=1in]{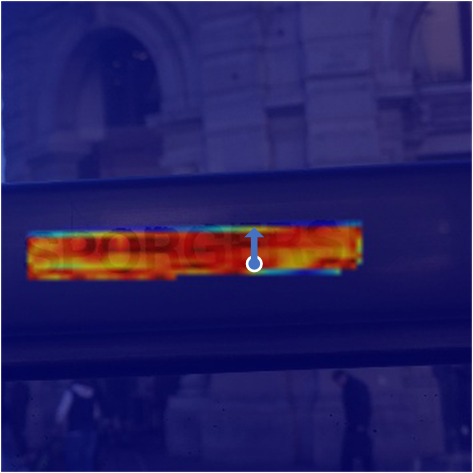}
			%\caption{fig2}
		\end{minipage}%
	}%
	\subfigure[bottom channel]{
		\begin{minipage}[t]{0.3\linewidth}
			\centering
			\includegraphics[width=1in, height=1in]{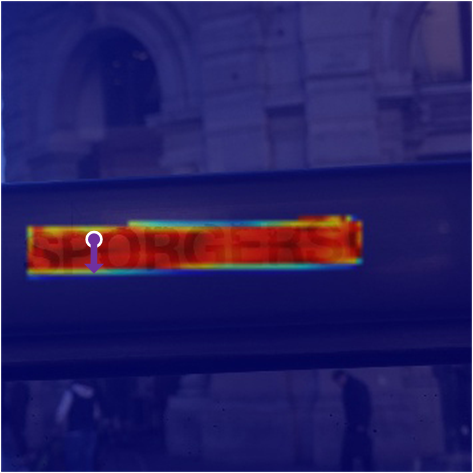}
			%\caption{fig2}
		\end{minipage}
	}%
	
	\subfigure[left channel]{
		\begin{minipage}[t]{0.3\linewidth}
			\centering
			\includegraphics[width=1in, height=1in]{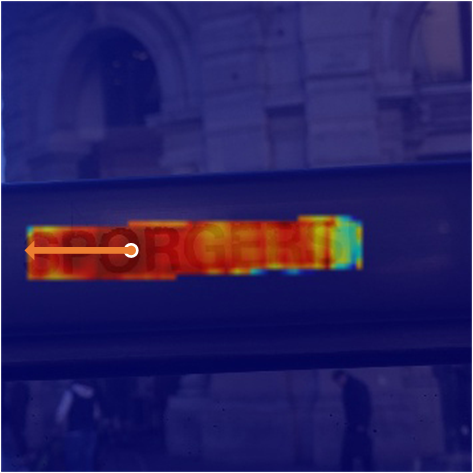}
			%\caption{fig1}
		\end{minipage}%
	}%
	\subfigure[right channel]{
		\begin{minipage}[t]{0.3\linewidth}
			\centering
			\includegraphics[width=1in, height=1in]{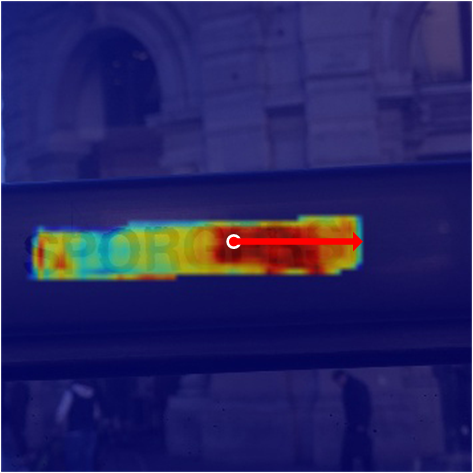}
			%\caption{fig2}
		\end{minipage}%
	}%
	\subfigure[angle channel]{
		\begin{minipage}[t]{0.3\linewidth}
			\centering
			\includegraphics[width=1in, height=1in]{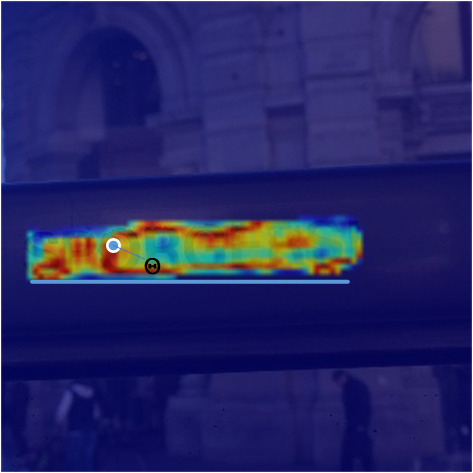}
			%\caption{fig2}
		\end{minipage}
	}%
	\caption{(a) is a text instance and (b)-(f) are the confidence (the accuracy of prediction) visualization maps about the distance to the four boundaries and rotation angle respectively in different locations. Red pixels mean higher confidence and blue is lower. We can observe that: (1) The most suitable features of different components are distributed in different locations; (2) The most appropriate feature is not always located at the location close to the predicted target (especially for top and bottom).}
	\label{figure2}
\end{figure}

Different from using the features from a single location, Lyu et al \cite{lyu2018multi} uses the features close to the targets. It predicts the four corners of bounding box separately and then combines them to form the final bounding box. The position of each corner is predicted by the feature close to itself. However, As shown in Fig. \ref{figure2} (a), the most suitable feature to predict the top boundary appears in the lower half of the text area, which means that the most suitable features are not always distributed near the predicted target. It’s also worth noting that the most suitable features to predict the left boundary and the right boundary surely locate in different positions obviously in Fig. \ref{figure2} (d) and Fig. \ref{figure2} (e), which further demonstrates the aforementioned irrationality of using features from one single location to predict all components of a bounding box.

\begin{figure}[htbp]
	\centering
	\subfigure[statistics example]{
		\begin{minipage}[t]{0.33\linewidth}
			\centering
			\includegraphics[width=1in]{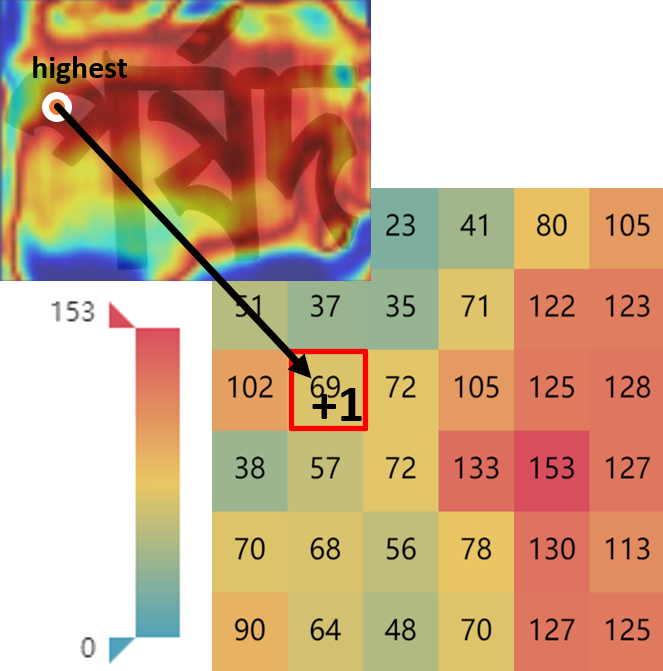}
		\end{minipage}%
	}%
	\subfigure[top channel]{
		\begin{minipage}[t]{0.33\linewidth}
			\centering
			\includegraphics[width=1in]{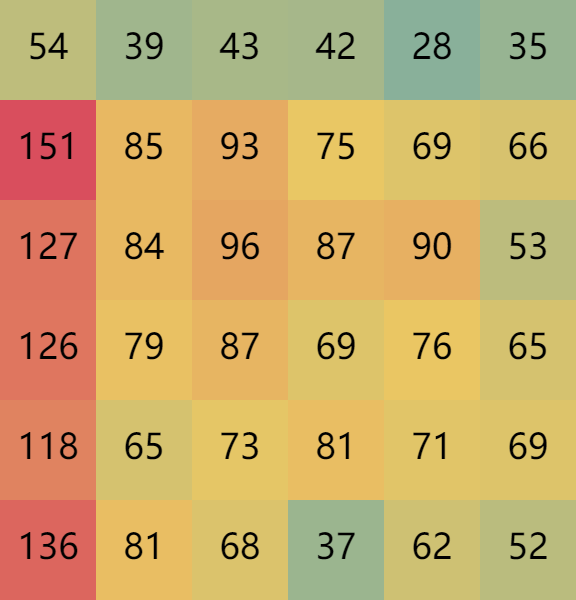}
			%\caption{fig1}
		\end{minipage}%
	}%
	\subfigure[bottom channel]{
		\begin{minipage}[t]{0.33\linewidth}
			\centering
			\includegraphics[width=1in]{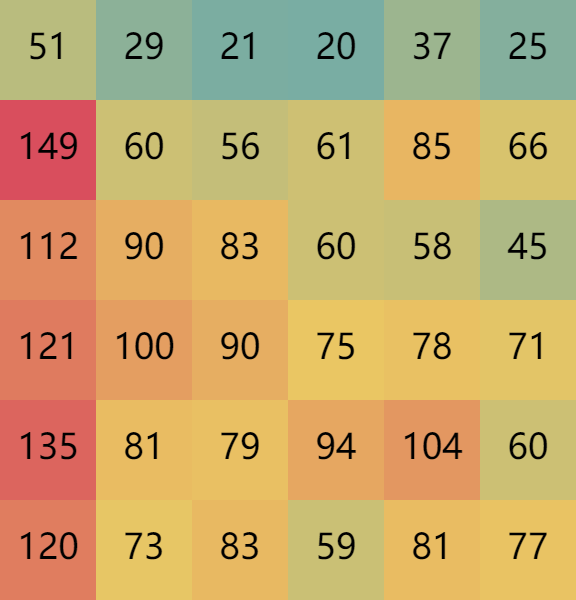}
			%\caption{fig2}
		\end{minipage}%
	}%
	
	\subfigure[left channel]{
		\begin{minipage}[t]{0.33\linewidth}
			\centering
			\includegraphics[width=1in]{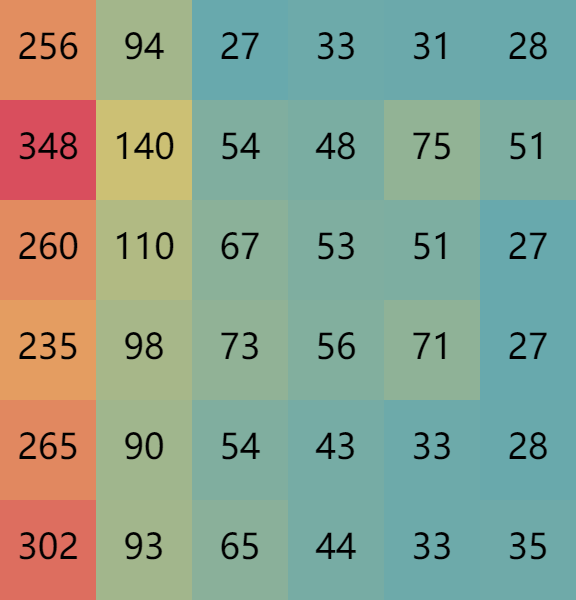}
			%\caption{fig2}
		\end{minipage}
	}%
	\subfigure[right channel]{
		\begin{minipage}[t]{0.33\linewidth}
			\centering
			\includegraphics[width=1in]{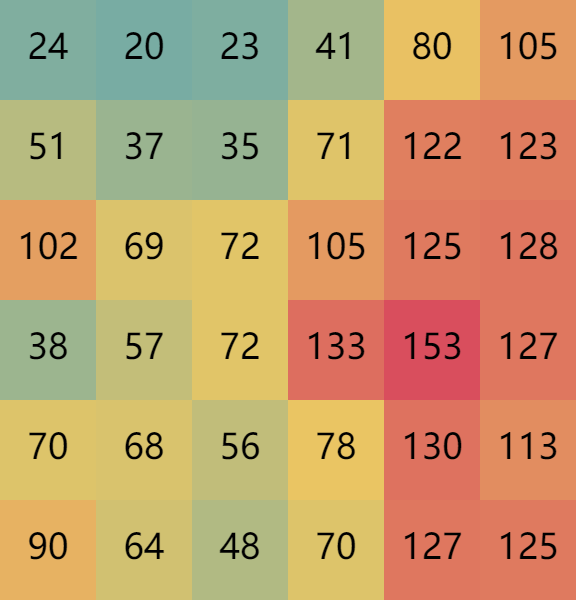}
			%\caption{fig1}
		\end{minipage}%
	}%
	\subfigure[angle channel]{
		\begin{minipage}[t]{0.33\linewidth}
			\centering
			\includegraphics[width=1in]{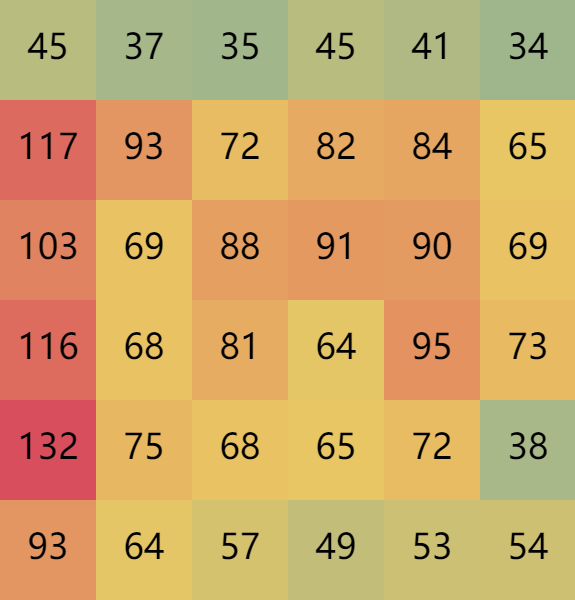}
			%\caption{fig2}
		\end{minipage}%
	}%
	\caption{Most suitable feature distribution statistics results. (a) is an example shows how we conducted statistics. (b)-(f) are the statistics results on corresponding channels. After resizing all text instances to 32$\times$32, we counted the distributions of the features with the highest confidence and visualized them. We can find the most suitable features don't have a stable distribution law. Therefore, it is of much necessity to propose a learnable feature selection way to flexibly select the suitable features.}
	\label{statistics}
\end{figure}

In order to further explore the relationship between the distribution of the most suitable features and the predicted target, we conducted statistics on the location distribution of the most suitable features. The text instances are from the validation set of ICDAR2017-MLT and the results are shown in Fig. \ref{statistics}. The numbers in the square represent the amounts of text instances whose most appropriate features are located at these positions. From Fig. \ref{statistics}, we can find that the most suitable features of different components are in different locations, and they are not always close to the predicted components, especially for top and bottom channels. These results indicate that selecting features in a single location to predict all components and selecting features in some certain regions to predict corresponding components are unreasonable. More importantly, it’s hard for us to summarize reasonable inherent law between feature location and predicted components in these distribution maps. Therefore, we propose to select the most suitable feature for each component by a learnable way.

In this paper, we propose a Location-Aware feature Selection text detection Network (LASNet) to separately learn the most appropriate locations of features for every component of the bounding box, and to improve the detection accuracy accordingly. We define the location’s confidence score to indicate whether its corresponding feature is suitable to predict the target component or not. Our LASNet first predicts the confidence maps of all five components of the bounding box separately. Then, we design a Location-Aware Feature Selection (LAFS) mechanism to select the components with the highest confidence score to form a more accurate bounding box for a text instance. Specifically, LAFS finds the top-$K$ features for each component and fuses them to get the best bounding box according to their confidence scores. By considering that the features from different locations fit with different components, LAFS lets network itself to learn the most suitable features’ distributions rather than fix features from one location or fix them in certain regions by human priori, which can increase the selection flexibility of each component prediction and thus promote the performance of the algorithm. After adding LAFS, LASNet has a significant accuracy improvement without efficiency loss.

The contributions of this paper are three-fold: (1)We find that there is no obvious regularity in the distribution of the suitable features of the predicted bounding box's components. And it's not accurate enough to use features from a single location or neighbor areas for bounding box prediction. (2) We propose an LASNet to learn the most appropriate locations of features for every component, and we design an LAFS mechanism to select and fuse the best components according to the confidence score, forming a more accurate bounding box result. (3) The proposed LASNet has reached the state-of-the-art performance, i.e. Hmean of 87.4\%, 69.6\% and 83.7\% on ICDAR2015, ICDAR2017-MLT and MSRA-TD500 respectively, outperforming all other regression-based detectors.

\section{Related Work}
% It is not a good choice to use the general object detection method to detect the text directly. Scene text has its own unique characteristics:
% (1) Most of the text exists in the form of long rectangle, which challenges the size of the feature map's receptive field of the general object detection networks. 
% (2) Tiny intervals are very common in natural scene text. Small space is easily overlooked after down-sampling in neural networks.
% (3) The feature of texts is sparse, which is not as obvious as the closed edge contour of the general object detection targets.
% (4) The spacing between words and sentences and the direction of the text challenge the text detection network. 
% Therefore, it is necessary to improve the general object detection methods to make them suitable for the task of text detection.

Driven by the deep neural network and large-scale datasets, scene text detection method has made great progress in the past few years. The mainstream text detection methods based on deep learning can be roughly divided into segmentation-based text detection and regression-based detection. Regression-based methods can be divided into indirect regression-based methods and direct Regression-based methods according to whether the anchor is used or not.

\subsection{Segmentation-Based Text Detection}
Most of the segmentation-based text detection methods improve FCN \cite{long2015fully} and Mask R-CNN \cite{he2017mask} on the task of text detection. PixelLink \cite{deng2018pixellink} uses instance segmentation to determine whether the pixels in different directions are in a text region. Then, Textsnake \cite{long2018textsnake} uses a number of disks with different sizes and directions to cover the annotation text, which can be used to detect arbitrary shape text. To distinguish adjacent text instances, PSENet \cite{li2018shape} uses the method of progressive scale expansion, and LSAE \cite{tian2019learning} regards text detection as instance segmentation and map image pixels into embedded feature space. Different from these methods, CRAFT \cite{baek2019character} first detects a single character (character region score) and a connection relationship (affinity score) between characters, followed by determining the final text line according to the connectivity between characters. Segmentation-based methods are faced with an essential problem, which is how to determine the scope of text instances. As a result, they often require a complex post-processing process and have degraded efficiency.

\subsection{Indirect Regression-Based Text Detection}
Indirect regression-based methods mostly draw lessons from general object detection networks, such as Faster R-CNN \cite{ren2015faster} and SSD \cite{liu2016ssd}. They generate text bounding boxes via predicting the bounding box offsets from anchors. For example, SLPR \cite{zhu2018sliding} adds coordinate regression of intersection point of horizontal and vertical uniform sliding line and text polygon on the traditional object detection framework Faster R-CNN. It realizes text detection of arbitrary shape. As two improvement methods of SSD in text detection, TextBoxes \cite{liao2017textboxes} modifies the shape of anchor and convolution kernel of SSD to make it more suitable for text line detection. Seglink \cite{shi2017detecting} no longer detects the entire text line at one time. It first detects text segments, and then connects segments together to get the final bounding box. Besides, RRD \cite{liao2018rotation} abandons the way of sharing feature map of classification and regression task which were used in previous detection framework, but adopts the way of independent feature extraction. Despite their comparable performance, indirect regression-based methods have to design anchor in advance, which is less robust to the changes in datasets and text scales.

\subsection{Direct Regression-Based Text Detection}
Direct regression-based approaches perform boundary regression by predicting the offsets from a given point. Among them, EAST \cite{EAST} uses FCN \cite{long2015fully} to predict text score map and combines bounding box predictions in text area to determine bounding boxes. TextMountain \cite{zhu2018textmountain} predicts text center-border probability (TCBP) and text center-direction (TCD) to easily separate text instances. In TCBP, mountaintop means the center of text instance and mountain foot means the border. An anchor-free RPN was proposed in AF-RPN \cite{zhong2019anchor}, which aims to replace the original anchor-based RPN in the Faster R-CNN framework to address the problems caused by the anchor. Direct regression-based methods are characterized by simple network structure and high detection efficiency. Nevertheless, they still have room for improvement in detection accuracy considering the hardness of getting the exact bounding box boundaries in existing methods.

To improve the bounding box prediction accuracy and keep the efficiency advantage of the direct regression-based text detectors, we propose a novel Location-Aware feature Selection text detection Network (LASNet) in this work. In order to effectively find the most suitable features for different components of the bounding box, our network learns the confidence of different locations' features for components prediction, selecting the most suitable features to form the superb feature combinations rather than bind them into the same location compulsively or just use the features closed to the predicted targets like most of other methods. The experimental results demonstrate that the LASNet not only keep the efficiency advantage of direct regression-based methods, but also achieved excellent accuracy, which is competitive with segmentation-based methods.

\begin{figure*}[htb]
	
	\centering
	\includegraphics[scale=0.25]{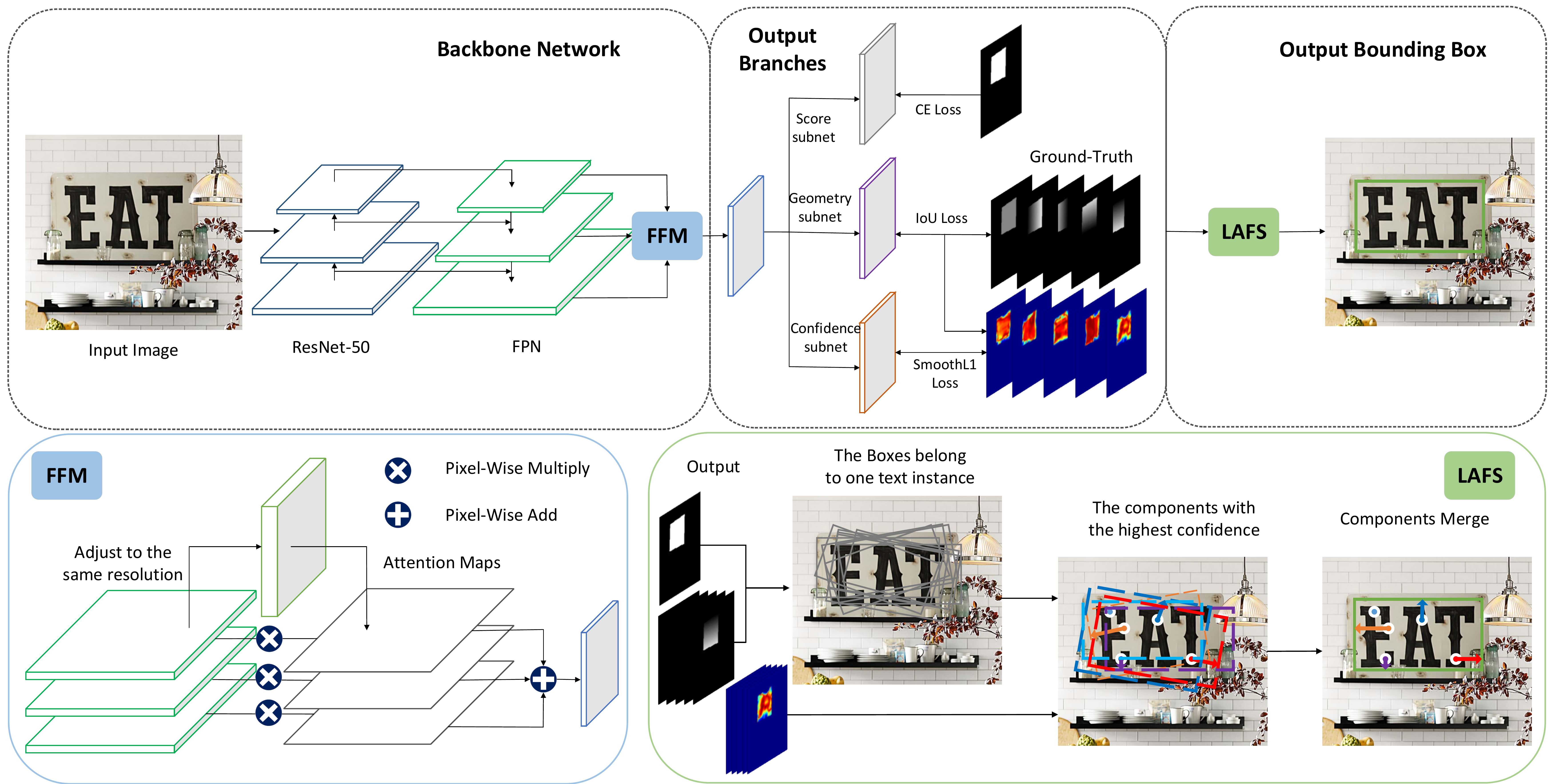}
	\caption{The architecture of LASNet. LASNet uses a Feature Pyramid Network \cite{long2015fully} backbone on top of a feed-forward ResNet \cite{he2016deep} architecture. The FPN outputs from different stages will be interpolated to the same resolution. In Feature Fusion Module (FFM), they are fused to one feature map with more suitable feature map by attention mechanism. Then there are three output branches, score, geometry and confidence. At the time of inference, proposed Location-Aware Feature Selection (LAFS) mechanism selects the feature with high confidence for each component and uses the most suitable feature combination to predict the final bounding box.}
	\label{Network}
	% \begin{minipage}[t]{1.0\textwidth}
	% \centering
	% \includegraphics[scale=0.5]{NetworkV4.png}
	% \end{minipage}
	
	% \begin{minipage}[t]{1.0\textwidth}
	% \centering
	% \includegraphics[scale=0.6]{av2.png}
	% \end{minipage}
\end{figure*}

\section{Our Method}

\subsection{Network Architecture}

\subsubsection{Network Architecture} Fig. \ref{Network} illustrates the architecture of the LASNet. First, we extract feature from multiple intermediate layers of ResNet-50 \cite{he2016deep}. Next comes an FPN \cite{lin2017feature}, by which we can get feature maps of different scales. Then it is a Feature Fusion Module (FFM). It can selectively fuse different stages’ features. For the output branches, apart from 1-channel classification score map and 5-channel geometry map, we especially predict a 5-channel confidence map used in LAFS at inference stage. Classification score map is a shrunk text foreground mask for the region of text instance, confirming that which features’ predictions are relatively accurate. Geometry map, containing 4 distances d$_t$, d$_l$, d$_b$, d$_r$ from the point to the box boundaries and a rotation angle $\theta$, is used for enclosing the word from the view of each pixel. In this work, the confidence map plays the key role in the proposed LAFS. Its channel number is the same as geometry map and it indicates the accuracy of the corresponding regression values in geometry map.

\subsubsection{Feature Fusion Module} As we all know, the abundance of feature map information is very helpful to improve the prediction accuracy of the network. We design a module called Feature Fusing Module (FFM). FFM uses the attention mechanism. Specifically, it predicts an attention map for each stage (three stages in this work) and selectively weights and fuses features from different stages to control their impact on the final prediction.

\subsection{Location-Aware Feature Selection}

To select the most suitable feature, we predict an extra confidence branch, whose five channels respectively correspond to the five channels of geometry branch. The meaning of confidence is the accuracy of the corresponding geometry branch's output. In the inference stage, we will disassemble five kinds of components of the bounding boxes which predict the same text instance and then select the components with higher confidences and finally reassemble the best 5 components into the best bounding box. The specific algorithm flow is shown in Algorithm \ref{LAFS}. First of all, we determine which locations in the feature map predict the same text instance. Then, we rank all the components of the bounding boxes according to their confidence. Each kind of component selects the top-$K$ confidence prediction results, $5\times K$ prediction results in all. Then the confidence is used as weight to fuse the $K$ prediction results of each component category. Finally, we combine the five fused results to form the final bounding box.

\begin{algorithm}
	\caption{The specific algorithm flow of Location-Aware Feature Selection}
	\label{LAFS}
	\begin{algorithmic}[1]
		\REQUIRE
		$B = [b_1,...,b_N]$, is $N \times 5$ matrix of initial detection boxes. $b_n = [d_t,d_b,d_l,d_r,\theta] \in B$. $S = [b_s]$, which is the collection of bounding boxes predicting the same text instance with $b_s$ (a random selected bounding box). $K$ means that we choose top-$K$ components. $U = []$ is the union of each kind of component with top-$K$ confidence. $U.clear()$ means empty $U$ to load next kind of component.
		\ENSURE
		$Box$ is the final output box of the text instance which $S$ stands for.
		\FOR{each $n \in [1,N]$}
		\IF{$IoU(b_s, b_n) > threshold$}
		\STATE $S \leftarrow S \cup b_n$
		\ENDIF
		\ENDFOR
		\FOR{each $comp \in [\theta,d_t,d_b,d_l,d_r]$}
		\FOR{$k=1$; $k<=K$; $k++$ }
		\STATE $idx \leftarrow max(S.comp.conf)$
		\STATE $U \leftarrow U \cup S[idx].comp$
		\STATE $S[idx].comp.conf \leftarrow 0$
		\ENDFOR
		\STATE $Box.comp \leftarrow merge(U)$
		\STATE $U.clear()$
		\ENDFOR
		\RETURN $Box$
	\end{algorithmic}
\end{algorithm}

Confidence map is defined as the confidence coefficient of each prediction of a text instance. In Eq. (\ref{conf}). $T$ means the text region. For a feature point $t_i$, its confidence $conf_i$ is defined as follows:

\begin{equation}
\begin{split}
\label{conf}
conf_i &=
\begin{cases}
1 - \dfrac{gap_i - min(Gap)}{\max(Gap) - min(Gap)} & if \quad t_i \in  T \\
0 & otherwise,
\end{cases}\\
gap_i &= abs(pred_i - label_i),
\end{split}
\end{equation}
where $gap_i$ is the gap between prediction and ground truth. As in the formula, we normalize the $conf_i$, so its range is [0, 1].

The specific details of function $merge()$ in algorithm flow chart is in Eq. (\ref{merge}):

\begin{equation}
merge(U) = \frac{\sum conf_{k} \times pred_{k}}{\sum conf_{k}}, 
\label{merge}
\end{equation}
where $k$ is in range $[1,K]$, $U$ is the union of confidence score and predicted components. And $pred$ means the position of a boundary or an angle of a bounding box. For the angle, $pred$ is same as $comp$ and we can get the final result by weighted sum in Eq. (\ref{merge}) directly. But for the boundary, we should firstly get the value of the $pred$ by the value of $comp$ and the position of the corresponding point. In addition, if the rotation angle of the bounding box is not zero, we also need to perform affine transformation.

In the process of inference, the key is how to determine which locations predict the same text instance. Common method is to instance segment the feature map \cite{deng2018pixellink,tian2019learning}. However, the instance segmentation often requires additional annotation information and network branches, which greatly increases the amount of computation. In this paper, we directly use the Intersection of Union (IoU) to deal with it. If the IoU of two locations’ predictions exceeds the threshold, we'll assume that they predict the same text instance.

\subsection{Loss Functions}
The loss function can be expressed as:

\begin{equation}
L_{sum} = L_{cls} + \gamma L_{geo} + \lambda L_{conf},
\end{equation}
where L$_{cls}$, $L_{geo}$ and L$_{conf}$ represent classification loss, regression loss and confidence loss respectively. $\gamma$ and $\lambda$ represent the hyper-parameters between different losses. According to the results of multiple experiments, the loss balance parameters can be set to $\gamma$ = 1 and $\lambda$ = 10.

\subsubsection{Loss for Classification Score Map}
In the classification branch, we use Dice Loss, which is usually used in segmentation tasks. Dice Loss can reduce the bad influence brought by the number unbalance between the negative and positive samples:

\begin{equation}
L_{cls} = 1 - (2 \times \frac{\sum \hat{S} \times S}{\sum \hat{S} + \sum S + \epsilon}),
\end{equation}
where $\hat{S}$ is the predicted value of the network, and $S$ represents the ground truth. $\epsilon$ is used to prevent dividing by zero.

\subsubsection{Loss for Geometries}
Geometries Loss has two parts, IoU Loss and Rotation Angle Loss. $\mu$ is a hyper-parameter. In our experiments $\mu$ = 10:

\begin{equation}
L_{geo} = L_{IoU} + \mu L_\theta,
\end{equation}
where L$_{IoU}$  represents IoU Loss:

\begin{equation}
L_{IoU} = - logIoU(\hat{R} \times R) = -log \frac{\hat{R} \cap R}{\hat{R} \cup R},
\end{equation}
where $\hat{R}$ and R represent the area of predicted bounding box and ground truth respectively, $\hat{R} \cap R$ means the intersection of areas.

\begin{equation}
\hat{R} \cup R = \hat{R} + R - \hat{R} \cap R .
\end{equation}

Rotation Angle Loss is:

\begin{equation}
L_\theta = 1 - cos(\hat{\theta} - \theta).
\end{equation}

We use cosine function to measure the angle gap between a prediction result and a ground truth.

\subsubsection{Loss for Confidence Map}
We select Smooth L1 loss as the Confidence Map Loss:

\begin{equation}
L_{conf} = SmoothL1(\hat{C}-C),
\end{equation}
where $\hat{C}$ is prediction and $C$ is the ground truth. 

\section{Experiments}
\subsection{Datasets}
The datasets we used for the experiments are briefly introduced below:

\noindent$\textbf{ICDAR2015(IC15)}$ \cite{karatzas2015icdar} comprises 1,000 training images and 500 testing images which is of different orientations. It is collected for the ICDAR 2015 Robust Reading Competition. The images are taken by Google Glasses without taking care of positioning, image quality, and viewpoint. IC15 only includes English text instances. The annotations are at the word level using quadrilateral boxes.

\noindent$\textbf{ICDAR2017-MLT(IC17)}$ \cite{nayef2017icdar2017} is a large-scale multi-lingual text dataset, which includes 7200 training images, 1800 validation images and 9000 test images with texts in 9 languages. The text regions in IC17 are also annotated by 4 vertices of the quadrangle.

\noindent$\textbf{MSRA-TD500}$ \cite{yao2012detecting} includes 500 images in total, 300 for training and 200 for testing, with text instances of different orientations. It is consisting of both English and Chinese text and annotated on text line level.

% \noindent$\textbf{SynthText}$ is a generated dataset in which word instances are placed in the natural scene image, taking the scene layout into account. SynthText consists of 800 thousand images, including about 8 million instances of compound words. Each text instance is annotated with its text string, word level, and character level bounding boxes.

\begin{figure*}[htbp]
	\begin{minipage}[t]{0.24\textwidth}
		\centering
		\includegraphics[width=4.3cm]{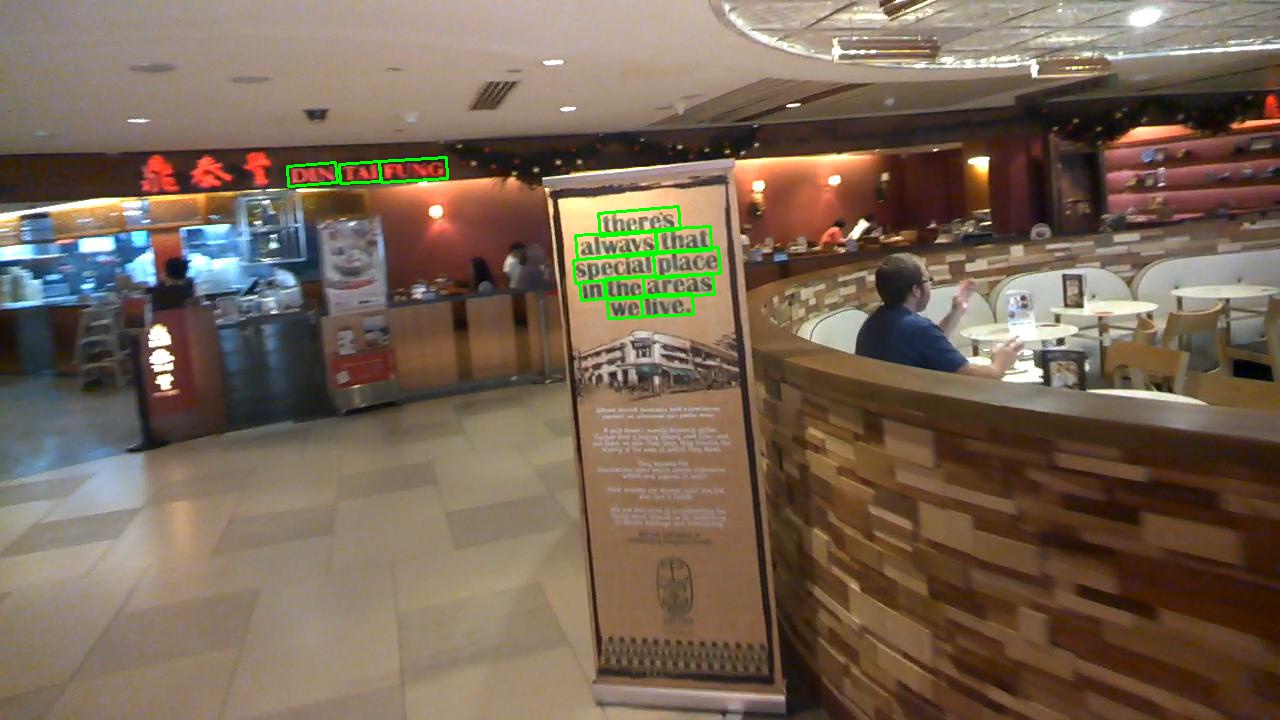}
	\end{minipage}
	\begin{minipage}[t]{0.24\textwidth}
		\centering
		\includegraphics[width=4.3cm]{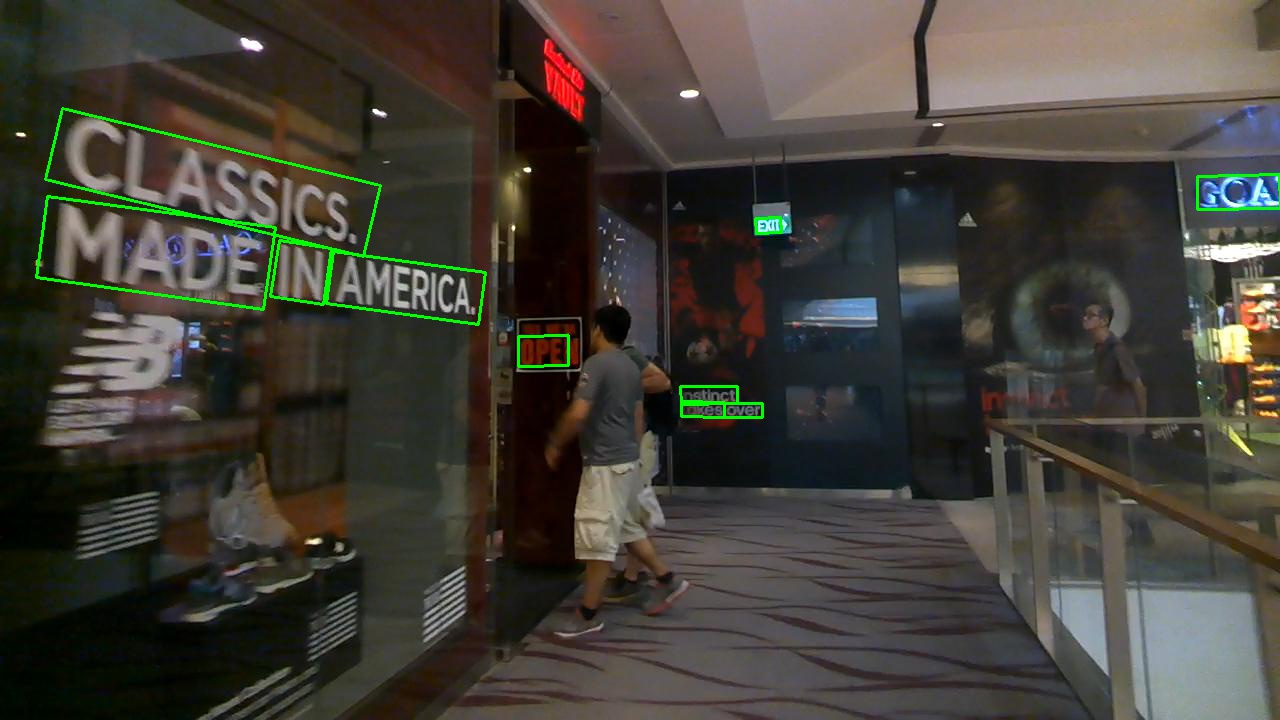}
	\end{minipage}
	\begin{minipage}[t]{0.24\textwidth}
		\centering
		\includegraphics[width=4.3cm]{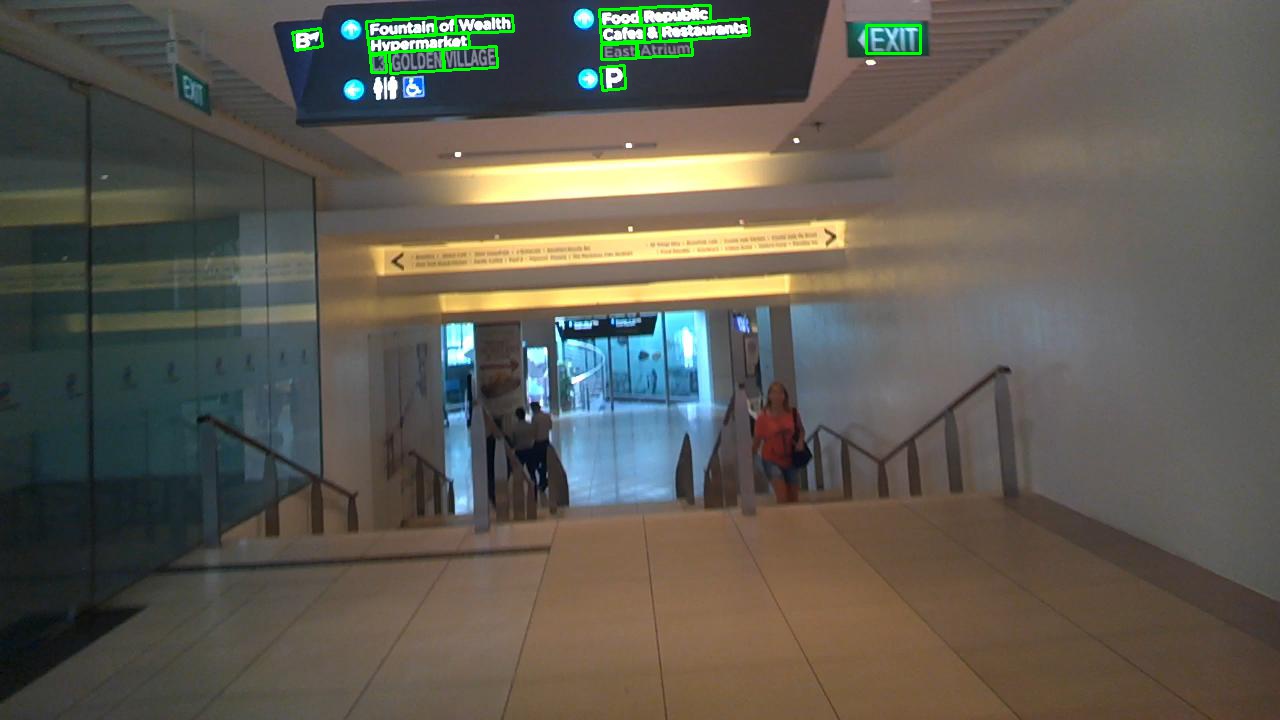}
	\end{minipage}
	\begin{minipage}[t]{0.24\textwidth}
		\centering
		\includegraphics[width=4.3cm]{2015_4.jpg}
	\end{minipage}

	\begin{minipage}[t]{0.24\textwidth}
		\centering
		\includegraphics[width=4.3cm]{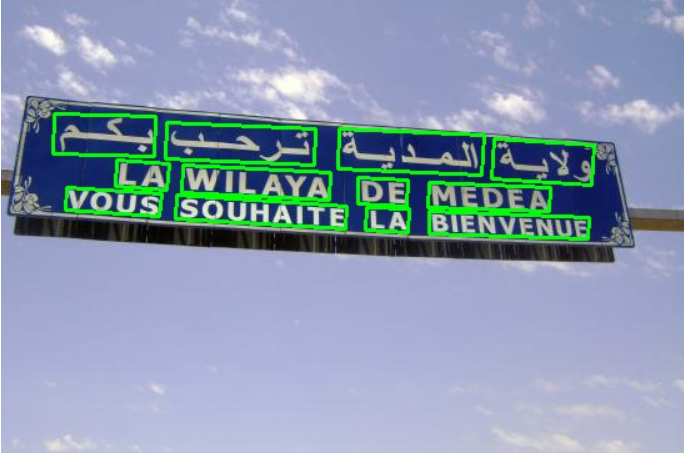}
	\end{minipage}
	\begin{minipage}[t]{0.24\textwidth}
		\centering
		\includegraphics[width=4.3cm]{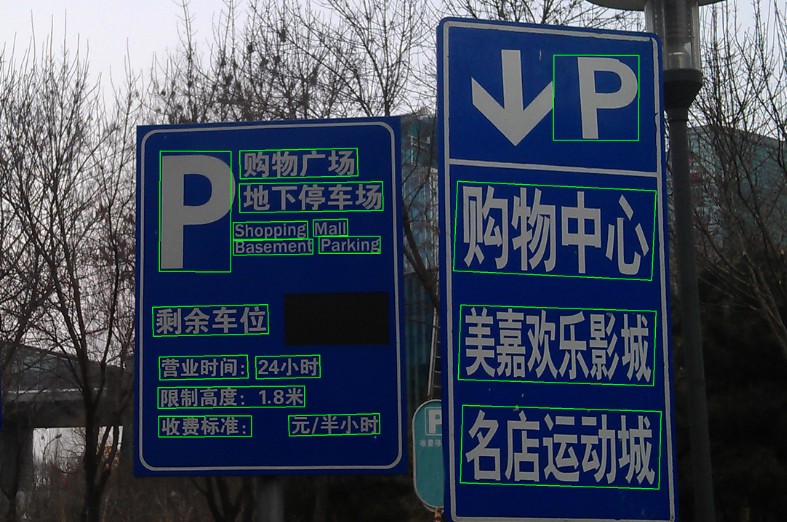}
	\end{minipage}
	\begin{minipage}[t]{0.24\textwidth}
		\centering
		\includegraphics[width=4.3cm]{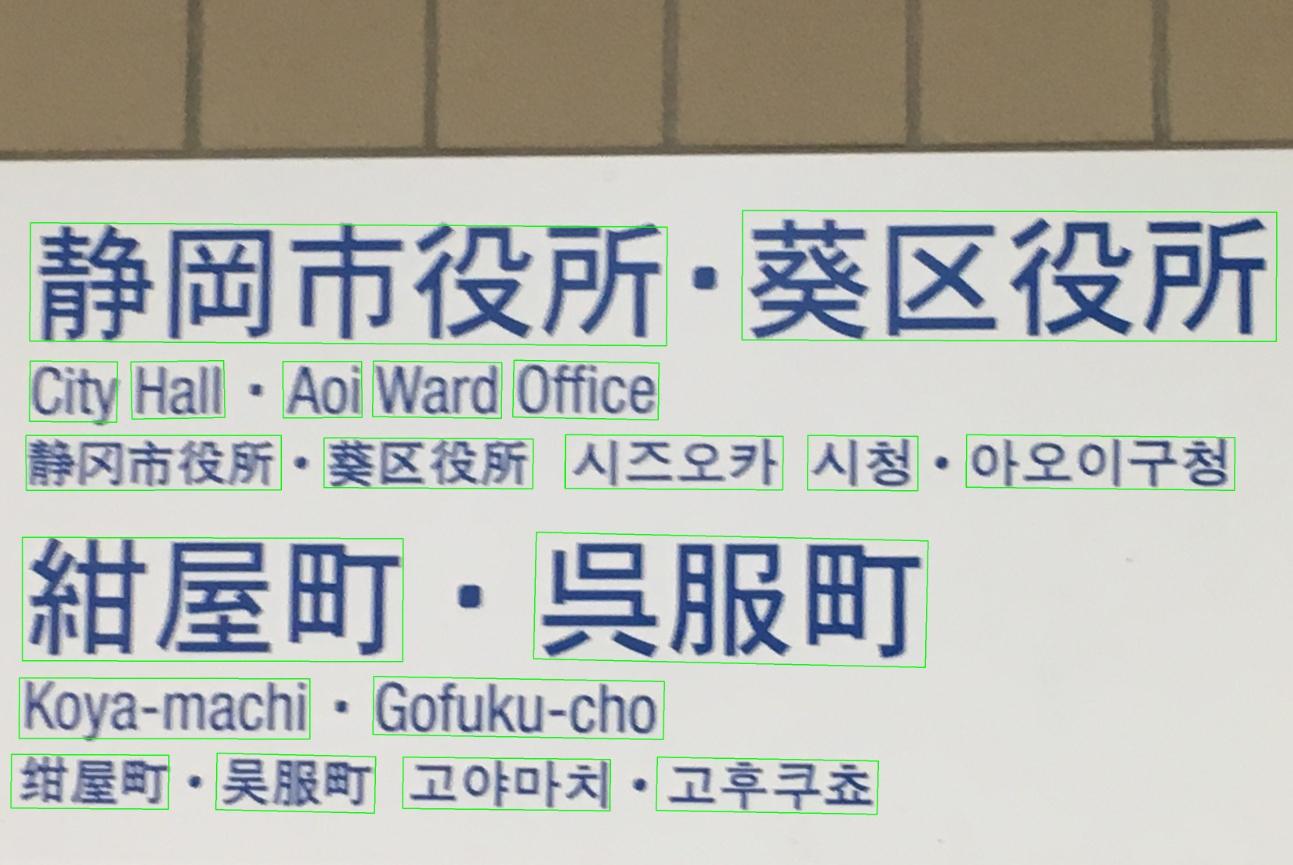}
	\end{minipage}
	\begin{minipage}[t]{0.24\textwidth}
		\centering
		\includegraphics[width=4.3cm]{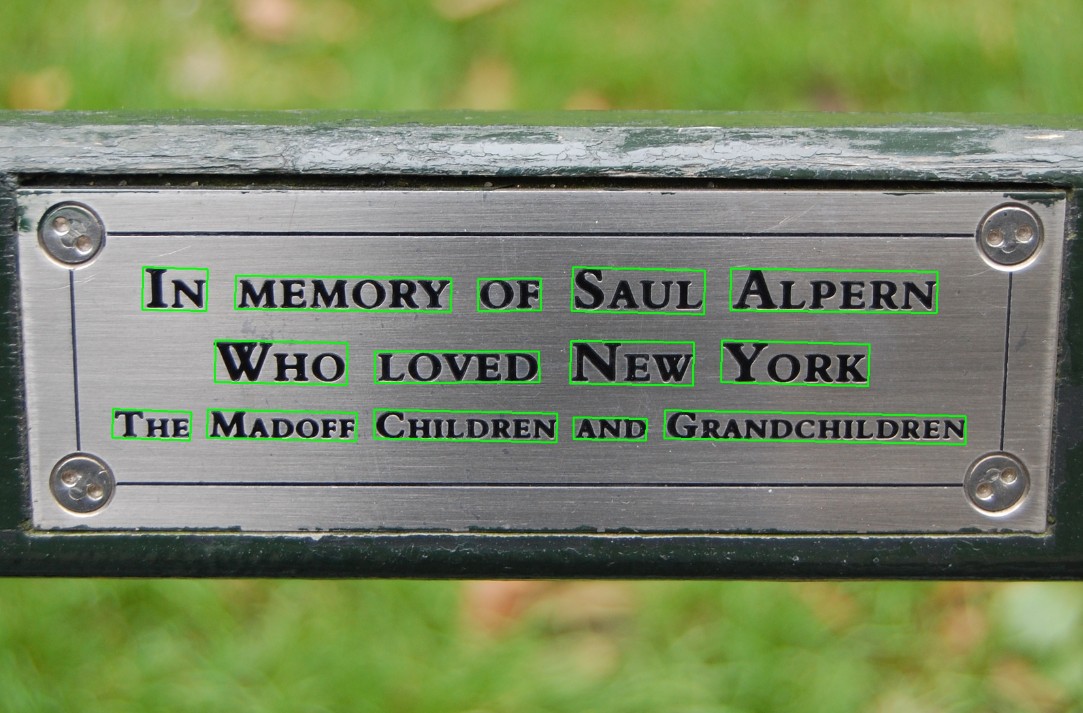}
	\end{minipage}

	\begin{minipage}[t]{0.24\textwidth}
		\centering
		\includegraphics[width=4.3cm]{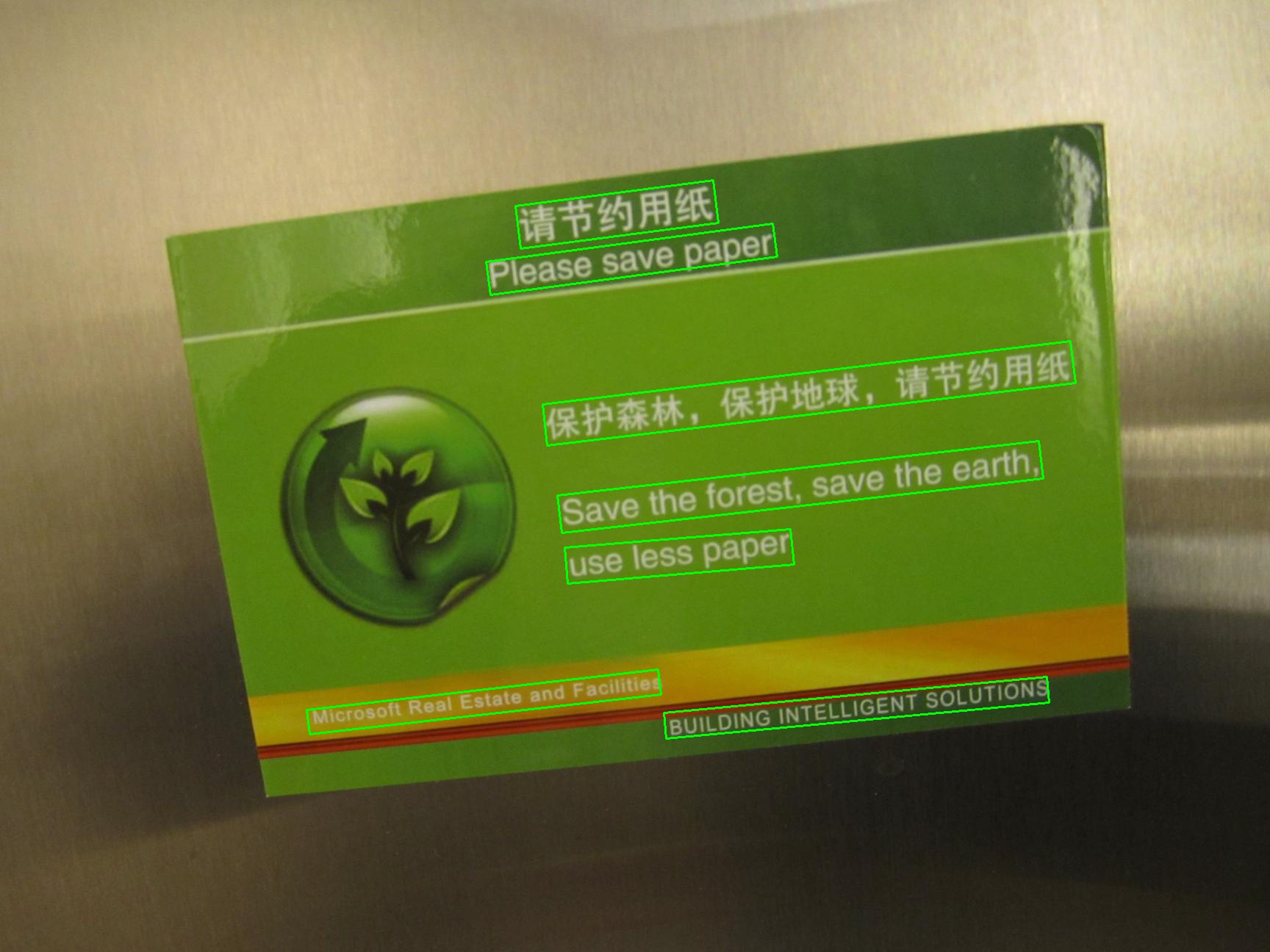}
	\end{minipage}
	\begin{minipage}[t]{0.24\textwidth}
		\centering
		\includegraphics[width=4.3cm]{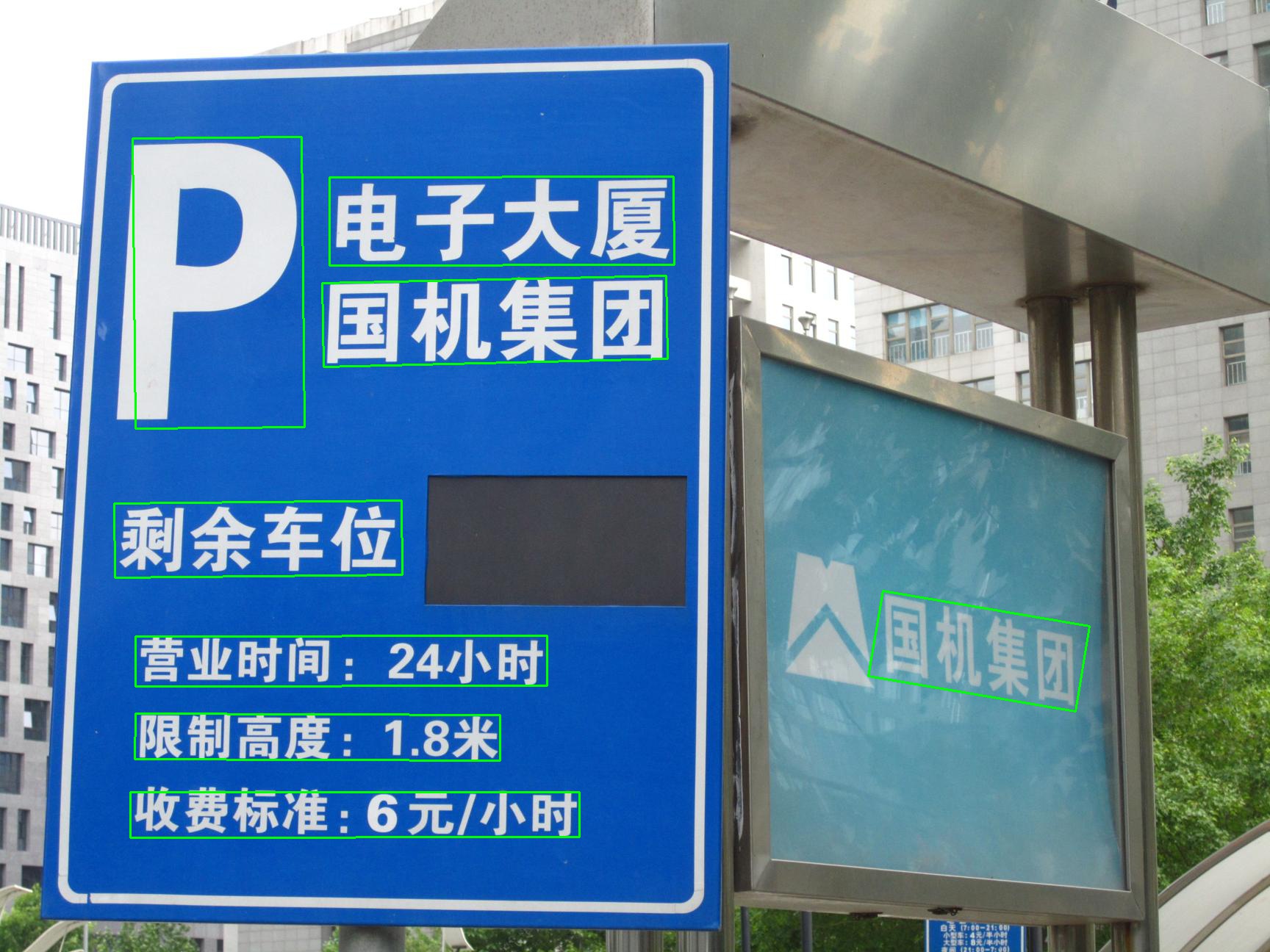}
	\end{minipage}
	\begin{minipage}[t]{0.24\textwidth}
		\centering
		\includegraphics[width=4.3cm]{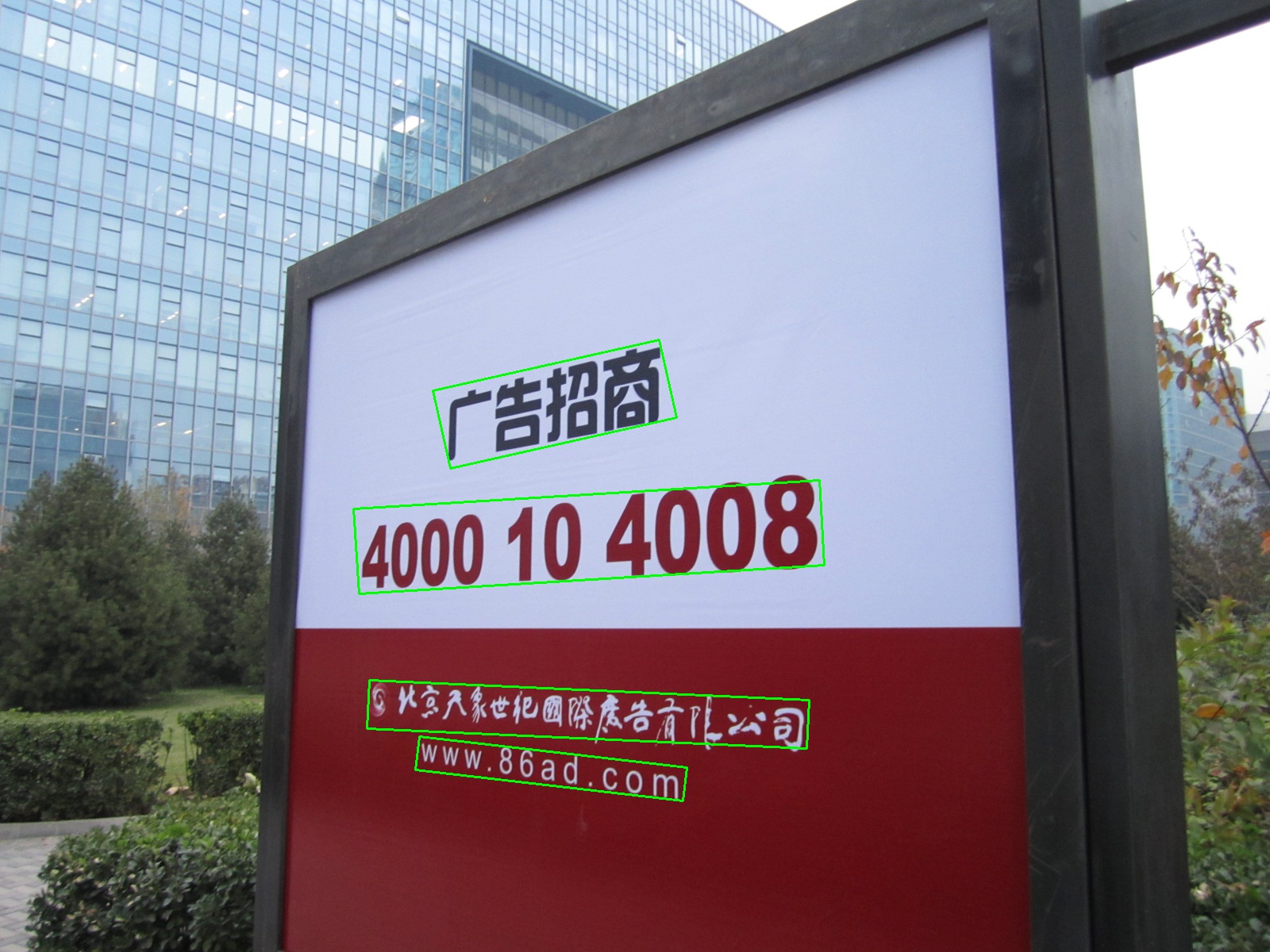}
	\end{minipage}
	\begin{minipage}[t]{0.24\textwidth}
		\centering
		\includegraphics[width=4.3cm]{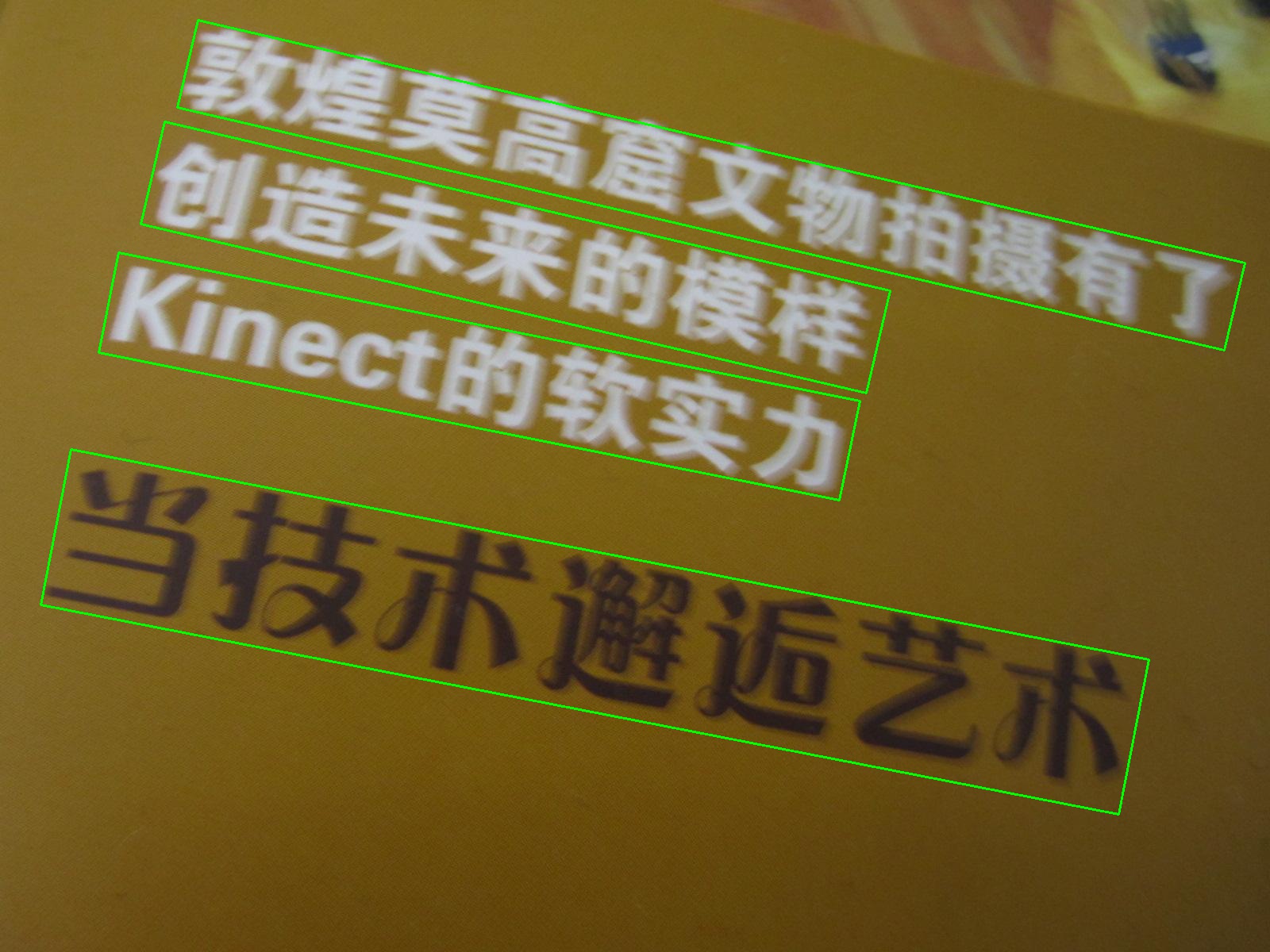}
	\end{minipage}
	\caption{Qualitative results on ICDAR2015, ICDAR2017-MLT and MSRA-TD500 by the proposed LASNet.}
	\label{Pictures}
\end{figure*}

\subsection{Training strategy}
\subsubsection{Experiment Settings} Our model is trained on double TITAN RTX GPUs by Adam optimizer \cite{kingma2014adam} with the batch size 32. Similar to EAST \cite{EAST}, we use the exponential decay learning rate strategy that the learning rate is multiplied by 0.94 for every 10,000 iterations, and the initial learning rate is set to 1.0$\times10^{-3}$ for all experiments. As the training sets of ICDAR2015 and MSRA-TD500 are too small, we directly pre-train our model on the union of MSRA-TD500, ICDAR2015, and ICDAR2017-MLT training datasets for 200,000 iterations, and then fine-tune the model with the training set of ICDAR2015, MSRA-TD500 and ICDAR2017 MLT respectively. ResNet-50 is taken as the backbone network, and the scale of training images is resized to 640.

\subsubsection{Data Augmentation} In our experiments, we follow the data augmentation of EAST \cite{EAST}. First, the sizes of images are randomly rescaled with ratio {0.5, 1.0, 2.0, 3.0} and  randomly rotate -25 to 25 degrees. Second, we crop images from the transformed images and the ratio of foreground to background is 3:1. Finally, the cropped images are resized to 640$\times$640 for training.

% \setlength{\tabcolsep}{6pt}
% \begin{table}[htbp]
% \begin{center}
% \caption{The efficiency comparison between segmentation-based methods ( $^\dag$ means segmentation-based method) and regression-based methods. Due to influences  of factors such as GPU and CPU, the FPS value is only for reference. But it can be seen that regression-based methods have a great advantage over segmentation-based methods in efficiency.}
% \label{efficiency}
% \begin{tabular}{lccc}
% \hline
% \textbf{Method}       & \textbf{Resolution} & \textbf{FPS} &\textbf{Dataset}               \\
% \hline
% \hline
% TextSnake$^\dag$\cite{long2018textsnake}    & 1280       & 1.1    &ICDAR2015           \\
% LOMO$^\dag$\cite{zhang2019look}         & 512        & 1.4    &SCUT-CTW1500          \\
% PixelLink$^\dag$\cite{deng2018pixellink}    & 1280       & 3.0    &ICDAR2015              \\
% TextField$^\dag$\cite{xu2019textfield}    & 1280       & 6.0    &ICDAR2015              \\
% CRAFT$^\dag$\cite{baek2019character}        & 960        & 8.6    &ICDAR2013    \\
% \hline
% RRD\cite{liao2018rotation}          & 1024       & 6.5    &ICDAR2015            \\
% TextBoxes++\cite{liao2018textboxes++}  & 1024       & 11.6   &ICDAR2015            \\
% EAST\cite{EAST}         & 720        & 13.2   &ICDAR2015            \\
% SRPN\cite{he2020realtime}         & 1280       & 16.5   &ICDAR2015            \\
% LASNet(ours)   & \textbf{2048}       & 9.3    &ICDAR2015            \\
% LASNet(ours)*   & 1280       & \textbf{17.4}   &ICDAR2015          \\
% \hline
% \end{tabular}
% \end{center}
% \end{table}

\subsection{Ablation Study}
% \usepackage{booktabs}
We conduct ablation experiments to verify the effectiveness of LAFS module. In this part, backbone is the ResNet-50\cite{he2016deep}. The dataset we use is the validation set of ICDAR2017-MLT. During the experiments, the longer side of the input images is set to 1024.

\setlength{\tabcolsep}{1.5mm}
\begin{table*}[htbp]
	\caption{We conducted the experiments on the validation set of IC17 to prove effectiveness of LAFS. "without LAFS" means use features from a single location to predict bounding box. "LAFS with constraint" means features selected to predict components must be close to the components,i.e.using feature closed to left boundary to predict the distance to the left boundary. "Gain" means the performance gap between LAFS}
	\label{compare}
	\begin{tabular}{c|ccc|ccc|ccc|c}
		\hline
		\textbf{IC17 MLT val}     & \multicolumn{3}{c|}{without LAFS} & \multicolumn{3}{c|}{LAFS with constraint} & \multicolumn{3}{c|}{LAFS} & Gain   \\ \hline
		IoU Threshold & Recall     & Precision    & Hmean      & Recall    & Precisioin  & Hmean & Recall    & Precisioin  & Hmean    & Hmean  \\ \hline
		0.5           & 63.8\%    & 92.5\%      & 75.5\%  & 64.1\%    & 91.7\%      & 75.5\%  & 64.5\%          & 92.7\%               & \textbf{76.1\%}   & 0.6\%/0.6\%   \\
		0.6           & 61.0\%    & 88.4\%      & 72.2\%  & 62.0\%    & 88.7\%      & 72.9\% & 62.6\%           & 90.0\%               & \textbf{73.8\%}   & 1.6\%/0.9\%   \\
		0.7           & 56.1\%    & 81.4\%      & 66.4\%  & 58.1\%    & 83.2\%      & 68.4\% & 58.6\%            & 84.2\%              &\textbf{69.1\%}  & 2.7\%/0.7\%   \\
		0.8           & 43.1\%    & 62.5\%      & 51.0\%  & 46.7\%    & 66.9\%      & 55.0\% & 46.7\%           & 67.1\%            &\textbf{55.1\%}  & 4.1\%/0.1\% \\ \hline
	\end{tabular}
\end{table*}

% \caption{Upper bound analysis of confidence via using ground-truth. We replace the predicted confidence with the ground-truth values, the results suggest there is still room for improvement in regressing confidence.}
% \label{upperboundanalysis}
% \begin{tabular}{c|ccc}
% \hline
% \textbf{IoU Threshold}        & \textbf{Recall} & \textbf{Precision} & \textbf{Hmean} \\
% \hline
% 0.5     & 69.7          & 87.3               & 77.5           \\
% 0.6           & 66.8           & 83.6               & 74.3          \\
% 0.7           & 62.2            & 77.8               & 69.1           \\
% 0.8    & 50.7           & 63.4               & 56.3           \\

\subsubsection{Assessment criteria of text detection} 
\begin{figure*}[htbp]
	\begin{center}
		\subfigure[]{
			\begin{minipage}[t]{0.24\linewidth}
				\centering
				\includegraphics[width=1.6in]{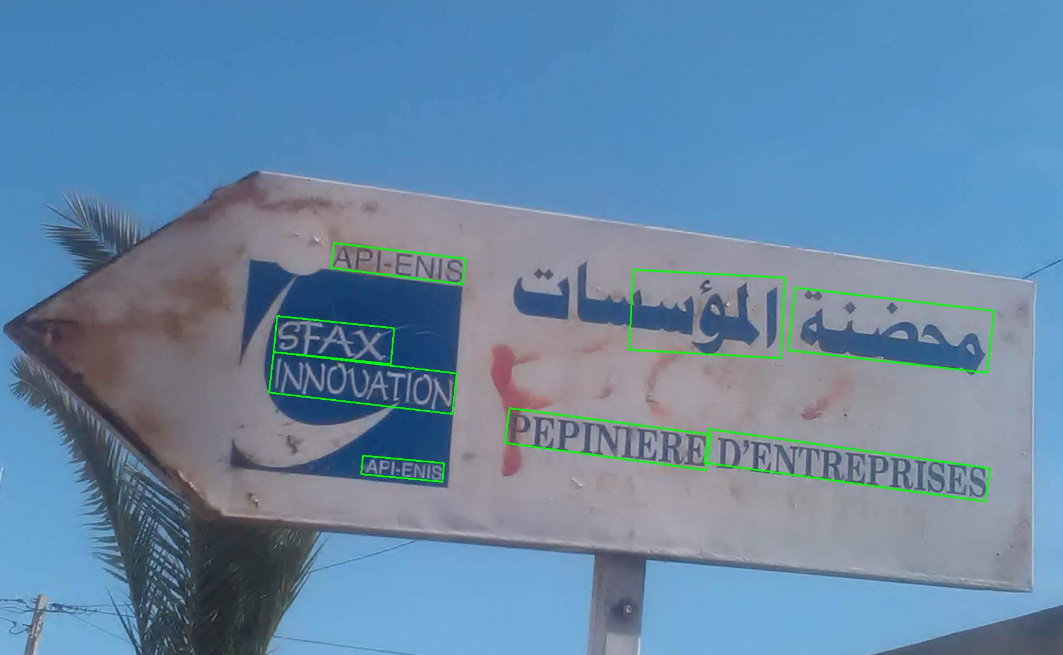}
				%\caption{Bounding box from EAST}
			\end{minipage}%
		}%
		\subfigure[]{
			\begin{minipage}[t]{0.24\linewidth}
				\centering
				\includegraphics[width=1.6in]{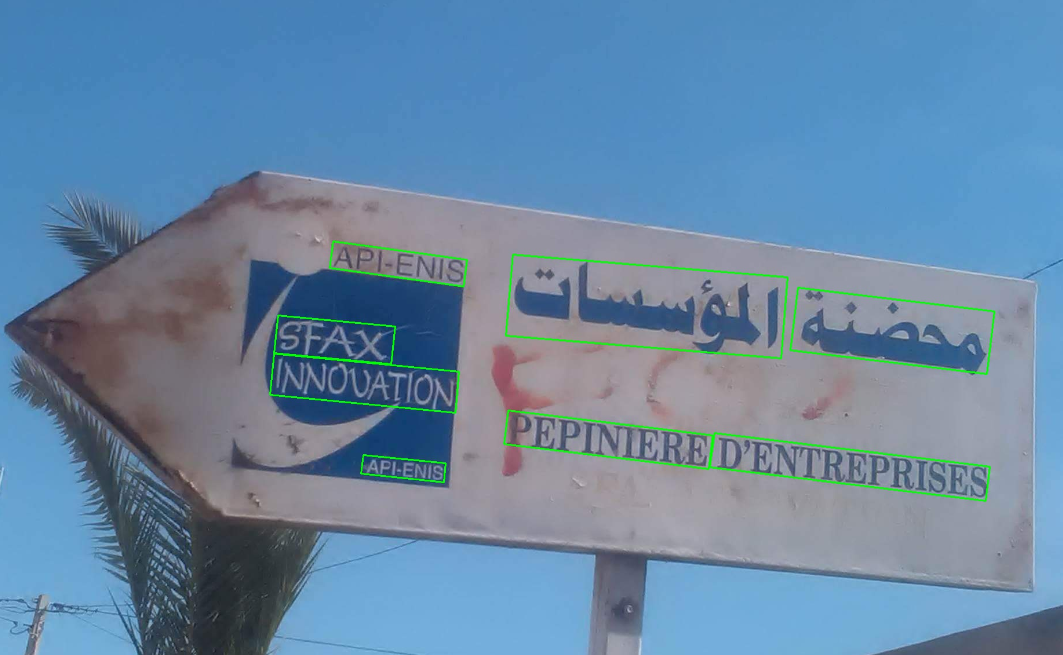}
				%\caption{Bounding box from LASNet}
			\end{minipage}%
		}%
		\subfigure[]{
			\begin{minipage}[t]{0.24\linewidth}
				\centering
				\includegraphics[width=1.6in]{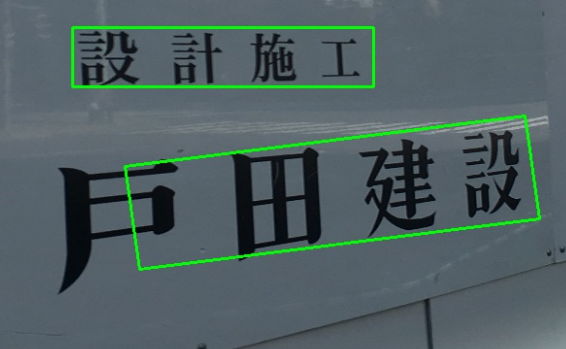}
				%\caption{Bounding box from EAST}
			\end{minipage}%
		}%
		\subfigure[]{
			\begin{minipage}[t]{0.24\linewidth}
				\centering
				\includegraphics[width=1.6in]{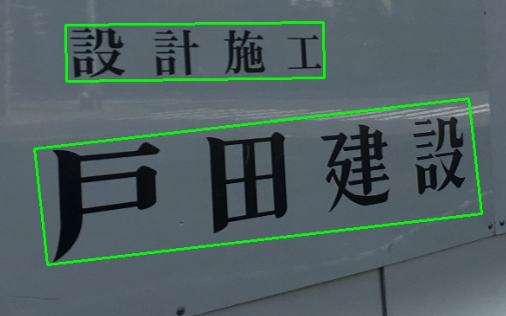}
				%\caption{Bounding box from LASNet}
			\end{minipage}%
		}%
		\centering
		\caption{(a) and (c) are the predicted bounding boxes if we only choose a single location features. (b) and (d) are the predicted bounding boxes after adding LAFS, choosing the best location's feature in the feature map for each component. (b) and (d) are more accurate than (a) and (c).}
		\label{criteria}
	\end{center}
	\setlength{\textfloatsep}{5pt}
\end{figure*}

In natural scene text detection, Recall, Precision and Hmean are usually used to measure the performance of the detectors. Hmean is calculated by the Recall and Precision, and the detailed process can be referred to the Wang et al \cite{wang2011end}. In the calculation of recall rate and accuracy, an important concept "match", namely the matched degree between the ground truth and its prediction, is involved. In the evaluation method of ICDAR dataset, if the Intersection of Union (IoU) between ground truth bounding box and prediction exceeds 0.5, they are supposed to be matched. Obviously, the higher the matching threshold, the more favorable for downstream works. As shown in the Fig. \ref{criteria} (a) and (c), the IoU of all text instances are higher than 0.5. But we can see that some of the predictions of the text instances are quite inaccurate for downstream works. Therefore, to better evaluate the accuracy of prediction,we also evaluated the higher threshold in the following experiments.

% This threshold setting is over-tolerant. For example, Fig. \ref{criteria} (a) and (c) miss part of the text, but the IoU between bounding boxes and ground truths exceeds 0.5, so they are supposed to be matched. However, these incomplete texts are hard to be applied in downstream works such as text recognition. If the IoU threshold is set to be higher, the recognition accuracy of the detected text boxes could be higher in a large part.

\subsubsection{Verification of LAFS} In order to show the effectiveness of the proposed LAFS, we conducted experiments on the IoU thresholds of 0.5, 0.6, 0.7 and 0.8 in the ablation experiments. In the process of inference, the value of $K$ is set to 1.

It can be concluded from Tab. \ref{compare} that our LAFS show greater performance advantages than the "without LAFS". Fig. \ref{criteria} provides the visualization examples of the improvement. From Fig. \ref{criteria} (a) and Fig. \ref{criteria} (c), we can get that when using single location features, some components of bounding boxes are predicted inaccurately. This is because the features used are not very suitable to predict these components, e.g. the distance of left boundary of the text instance in Fig. \ref{criteria} (a) and Fig. \ref{criteria} (c). However, after adding LAFS, we can use confidence map to find the locations of features which are suitable to predict these components. By combining the various components predicted by different features, we can get the final accurate bounding boxes as shown in Fig. \ref{criteria} (b) and Fig. \ref{criteria} (d). When the IoU threshold in the evaluation process raises, we find that LAFS has more significant accuracy improvement. That's because LAFS uses more appropriate features that make the prediction box more accurate. Under stricter IoU thresholds, more predicted text bounding boxes are matched with ground truth compared with using single location features, which leads to high Recall and Precision. When we set the IoU threshold as 0.8, compared with the "without LAFS", our LAFS can get 3.6\% improvement on Recall, 4.6\% improvement on Precision and 4.1\% improvement on Hmean. This result is a further evidence of the remarkable improvement in performance for bounding box prediction based on our LAFS. 

In order to compare the accuracy between the way of selecting features from the neighbor areas with the proposed LAFS, we conduct the "LAFS with constraint" experiments, in which we add constraint to the prediction of boundaries: only the features of the neighbor areas can be selected. We define the neighbor areas as where the distance from the predicted boundary less than 1 / 3 of the length of the adjacent edge in the text box. The experimental results are shown in the "LAFS with constraint" part of Tab. \ref{compare}, which indicate that all accuracy values are descended after adding the constraint of neighbor area. In other words, the proposed LAFS could achieve better bounding box prediction through selecting the best features from the whole map.

\setlength{\tabcolsep}{6pt}
\begin{table}[htbp]
	\begin{center}
		\caption{The selection of hyper-parameter $K$. We conducted the experiments on the validation set of ICDAR2017-MLT. The IoU threshold is 0.8.}
		\label{K}
		\begin{tabular}{l|ccc}
			\hline
			\textbf{$K$}                & \textbf{Recall}       & \textbf{Precision}       & \textbf{Hmean}       \\ 
			\hline
			\hline
			1       & 46.7\%& 67.1\% & 55.1\% \\
			2       & 47.3\% & 68.1\% & 55.9\% \\
			4       & 47.2\% & 68.1\% & 55.8\% \\
			6       & 47.1\% & 68.0\% & 55.7\% \\
			\hline
		\end{tabular}
	\end{center}
\end{table}

\subsubsection{Multi-group of Components} We consider the top-$K$ group of components, and then through the confidence of the components, weighted and integrated them, getting our final prediction box for the consideration of the robustness. We set the group number $K$ as a hyper-parameter, and Tab. \ref{K} shows the ablation experiment we conducted for the hyper-parameter $K$. If the group number is increased from 1 to 2, the performance may improve. But if we choose too many groups, the performance will decrease instead. We think this is because of the introduction of too many inaccurate prediction information. According to Tab. \ref{K}, in the latter part of the experiment, $K$ is set to 2.

\subsection{Experiments on Scene Text Benchmarks}

\begin{figure}[htb]
	\centering
	\includegraphics[scale=0.21]{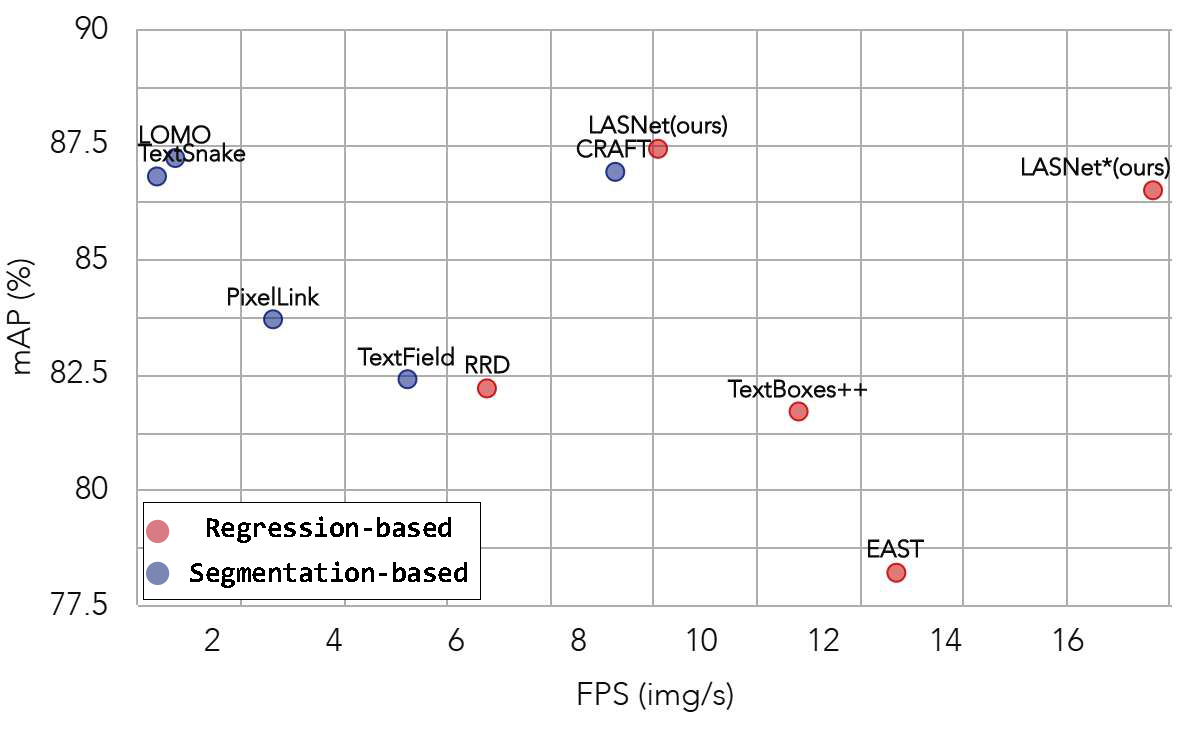}
	\caption{The efficiency comparison between segmentation-based methods (blue) and regression-based methods (red) on ICDAR2015. LASNet* means the length of text long side is 1280. Due to influences of factors such as GPU and CPU, the FPS value is only for reference. It can be seen that regression-based methods have a great advantage over segmentation-based methods in efficiency.}
	\label{efficiency}
\end{figure}

\subsubsection{Efficiency Comparison}
Segmentation-based methods sometimes have better performance than regression-based methods. However, in order to segment text instances which have small distances with each other and form integral bounding boxes, segmentation-based methods always have complex post process exerted on all pixels of text regions. In contrast, regression-based methods can exclude most of pixels by reasonable threshold and get integral bounding boxes by remaining pixels. As a result, the efficiency of segmentation-based methods is lower than regression-based methods. We compare the efficiency between segmentation-based methods and regression-based methods. As shown in Fig. \ref{efficiency}, The blue dots (segmentation-based methods) are located in the left side of the chart and the red dots (regression-based methods) are on the right almost. It is conspicuous that regression-based methods have huge advantages on efficiency. More importantly, our LASNet outperforms all the previous text detection methods on IC15, equipped with both high accuracy and efficiency. And our faster version can run up to two times faster than the fastest segmentation-based method while maintaining the competitive accuracy with it.

\setlength{\tabcolsep}{6pt}
\begin{table}[htbp]
	\begin{center}
		\caption{Quantitative results of different methods evaluated on ICDAR2015. $^\dag$ means segmentation-based method.}
		\label{IC15}
		\begin{tabular}{lccc}
			\hline
			\textbf{Method}        & \textbf{Recall} & \textbf{Precision} & \textbf{Hmean} \\
			\hline
			\hline
			TextSnake$^\dag$ \cite{long2018textsnake} & 80.4   &84.9               &86.8               \\
			PixelLink$^\dag$ \cite{deng2018pixellink}              & 82.0            & 85.5               & 83.7           \\
			FTSN$^\dag$ \cite{dai2018fused}                   & 80.0            & 88.6               & 84.1           \\
			TextMountain$^\dag$ \cite{zhu2018textmountain}     &84.2               &88.5               &86.3               \\
			LSAE$^\dag$ \cite{tian2019learning}                   & 85.0            & 88.3               & 86.6           \\
			IncepText$^\dag$ \cite{yang2018inceptext}             & 80.6            & 90.5               & 85.3           \\
			CRAFT$^\dag$ \cite{baek2019character}                 & 84.3            & 89.8               & 86.9           \\
			SPCNet$^\dag$ \cite{xie2019scene}                 & 85.8            & 88.7               & 87.2           \\
			PSENet$^\dag$ \cite{li2018shape}                & 85.2            & 89.3               & 87.2           \\
			LOMO$^\dag$ \cite{zhang2019look}                   & 83.5            &91.3              & 87.2           \\
			\hline
			EAST \cite{EAST}                   & 73.5            & 83.6               & 78.2           \\
			R2CNN\cite{jiang2017r2cnn}                             & 79.7           & 85.6                 & 82.5              \\
			CRPN \cite{deng2019detecting}                   & 80.7            & 88.8               & 84.5           \\
			SLPR \cite{zhu2018sliding}                   & 83.6            & 85.5               & 84.5           \\
			LASNet(ours)                                  & 84.0              &91.2                   &87.4\\
			\hline
		\end{tabular}
	\end{center}
\end{table}

\subsubsection{Accuracy Comparison: ICDAR2015} We evaluate the performance of our method on predicting multi-orientation natural scene text. Because ICDAR2015 has too few training images, we use both ICDAR2015 and ICDAR2017-MLT training sets to fine tune 50k steps. In the test process, we use single-scale test (length of text long side is 2048), which can achieve state-of-the-art performance (R: 84.0\%, P: 91.2\%, H: 87.4\%) comparing with other text detection methods, including both the segmentation-based methods and the regression-based methods as shown in Tab. \ref{IC15}. In ICDAR2015, text annotation is words level. The results show that our method can well solve the detection task of multi-orientation word level annotation text in natural scenes.

\subsubsection{Accuracy Comparison: ICDAR2017-MLT} We evaluate the performance of our method with more complex scenarios on ICDAR2017-MLT. We use the training set of ICDAR2017-MLT to fine tune 50k steps, and use a single-scale test (length of long side is 1920). Compared with ICDAR2015, the text instances of ICDAR2017-MLT have more diversity, larger scale changes and multiple languages. Our method can greatly improve the performance of regression-based detectors on this dataset as shown in Tab. \ref{IC17}. We achieve the performance of 69.6\% of Hmean, which is state-of-the-art in the regression-based methods. Due to the dramatic scale changes of text instances in IC17, regression-based methods are hard to determine the scope of the regression, which makes the inferior accuracy performance comparing with segmentation based methods. However, taking both accuracy and efficiency into account, the regression-based methods, especially our proposed LASNet, have greater potential.

%This indicates the promising potential of our proposed regression-based method for taking both accuracy and efficiency into account. But compared with the current segmentation-based methods, LASNet still has a gap. This is due to the particularity of dataset IC17, i.e. the text instances scale changes very dramatically.

\setlength{\tabcolsep}{6pt}
\begin{table}[htbp]
	\begin{center}
		\caption{Quantitative results of different methods evaluated on ICDAR2017-MLT. $^\dag$ means segmentation-based method.}
		\label{IC17}
		\begin{tabular}{lccc}
			\hline
			\textbf{Method}        & \textbf{Recall} & \textbf{Precision} & \textbf{Hmean} \\
			\hline
			\hline
			Lyu et al.$^\dag$ \cite{lyu2018multi}     & 56.6            & 83.8               & 66.8           \\
			LOMO$^\dag$ \cite{zhang2019look}            & 60.6            & 78.8               & 68.5           \\
			SPCNet$^\dag$ \cite{xie2019scene}          & 66.9       &73.4               &70.0                \\
			PSENet$^\dag$ \cite{li2018shape}         & 68.4            & 77.0               & 72.5           \\
			CRAFT$^\dag$ \cite{baek2019character}          & 68.2            & 80.6               & 73.9           \\
			Huang et al.$^\dag$ \cite{huang2019mask}                      &69.8           &80.0               &74.3           \\
			\hline
			He et al. \cite{he2018multi}       & 57.9            & 76.7               & 66.0           \\
			Border \cite{xue2018accurate}         & 62.1            & 77.7               & 69.0           \\
			LASNet(ours)   & 61.0            &81.1                  &69.6\\
			\hline
		\end{tabular}
	\end{center}
\end{table}

\subsubsection{Accuracy Comparison: MSRA-TD500} We evaluate the suitability of our method to long text on MSRA-TD500. Unlike ICDAR2015 and ICDAR2017-MLT, MSRA-TD500 annotations are text lines rather than words. We use MSRA-TD500 training set to fine tune 50K steps and single-scale test (length of long side is 1024) is used in the test process. As shown in Tab. \ref{TD500}, we achieve state-of-the-art performance (R: 80.7\%, P: 87.0\%, H: 83.7\%). The result also indicates that our method has a good performance in dealing with long text for MSRA-TD500 has a large quantity of long text instances.

\setlength{\tabcolsep}{6pt}
\begin{table}[htbp]
	\begin{center}
		\caption{Quantitative results of different methods evaluated on MSRA-TD500. $^\dag$ means segmentation-based method.}
		\label{TD500}
		\begin{tabular}{lccc}
			\hline
			\textbf{Method}        & \textbf{Recall} & \textbf{Precision} & \textbf{Hmean} \\
			\hline
			\hline
			Lyu et al.$^\dag$ \cite{lyu2018multi}     & 76.2           & 87.6               & 81.5           \\
			FTSN$^\dag$ \cite{dai2018fused}           & 77.1            & 87.6               & 82.0           \\
			LSAE$^\dag$ \cite{tian2019learning}           & 81.7            & 84.2               & 82.9           \\
			\hline
			Zhang et al. \cite{zhang2016multi}    & 67.0            & 83.0               & 74.0           \\
			Yao et al. \cite{yao2016scene}       & 75.3            & 76.5               & 75.9           \\
			EAST \cite{EAST}           & 67.4            & 87.3               & 76.1           \\
			SegLink \cite{shi2017detecting}        & 70.0            & 86.0               & 77.0           \\
			RRD \cite{liao2018rotation}            & 73.0            & 87.0               & 79.0           \\
			Border \cite{xue2018accurate}          & 77.4            & 83.0               & 80.1           \\
			ITN \cite{wang2018geometry}            & 72.3            &90.3              & 80.3           \\
			LASNet(ours)  &80.7            & 87.0               & 83.7 \\
			\hline
		\end{tabular}
	\end{center}
\end{table}

\subsection{Conclusion and Future Work}
%We propose a regression-based scene text detection method called location-Aware Feature Selective (LAFS).   From the ceiling analysis experiment, we can see that its potential is huge. We can continue to study in the design and the prediction of feature confidence. At the same time, we also think that the idea of feature suitability is not only effective in detection tasks, but also can be explored in other directions of computer vision, such as classification and segmentation tasks.

We propose a regression-based scene text detection method called Location-Aware feature Selection text detection Network (LASNet). In this work, we discover that one important reason of the inaccuracy of regression-based text detection methods is that they usually utilize a fixed feature selection way. To address this issue, we design LAFS mechanism to flexibly select the features. Our network separately learns the confidence of different locations’ features and selects the suitable features for the components of the bounding box by LAFS. We evaluate our method on ICDAR2015, ICDAR2017-MLT and MSRA-TD500, proving the performance improvement brought by our method. As for future work, in our opinion, LASNet is a beginning to consider feature selection in text detection tasks. Exploring more precise methods of feature selection may be a key research direction in the future.

{\small
\bibliographystyle{ieee_fullname}
\bibliography{egbib.bib}

\begin{thebibliography}{10}\itemsep=-1pt

\bibitem{baek2019character}
Youngmin Baek, Bado Lee, Dongyoon Han, Sangdoo Yun, and Hwalsuk Lee.
\newblock Character region awareness for text detection.
\newblock In {\em Proceedings of the IEEE Conference on Computer Vision and
  Pattern Recognition}, pages 9365--9374, 2019.

\bibitem{dai2018fused}
Yuchen Dai, Zheng Huang, Yuting Gao, Youxuan Xu, Kai Chen, Jie Guo, and Weidong
  Qiu.
\newblock Fused text segmentation networks for multi-oriented scene text
  detection.
\newblock In {\em 2018 24th International Conference on Pattern Recognition
  (ICPR)}, pages 3604--3609. IEEE, 2018.

\bibitem{deng2018pixellink}
Dan Deng, Haifeng Liu, Xuelong Li, and Deng Cai.
\newblock Pixellink: Detecting scene text via instance segmentation.
\newblock In {\em Thirty-second AAAI conference on artificial intelligence},
  2018.

\bibitem{deng2019detecting}
Linjie Deng, Yanxiang Gong, Yi Lin, Jingwen Shuai, Xiaoguang Tu, Yuefei Zhang,
  Zheng Ma, and Mei Xie.
\newblock Detecting multi-oriented text with corner-based region proposals.
\newblock {\em Neurocomputing}, 334:134--142, 2019.

\bibitem{he2017mask}
Kaiming He, Georgia Gkioxari, Piotr Doll{\'a}r, and Ross Girshick.
\newblock Mask r-cnn.
\newblock In {\em Proceedings of the IEEE international conference on computer
  vision}, pages 2961--2969, 2017.

\bibitem{he2016deep}
Kaiming He, Xiangyu Zhang, Shaoqing Ren, and Jian Sun.
\newblock Deep residual learning for image recognition.
\newblock In {\em Proceedings of the IEEE conference on computer vision and
  pattern recognition}, pages 770--778, 2016.

\bibitem{he2018multi}
Wenhao He, Xu-Yao Zhang, Fei Yin, and Cheng-Lin Liu.
\newblock Multi-oriented and multi-lingual scene text detection with direct
  regression.
\newblock {\em IEEE Transactions on Image Processing}, 27(11):5406--5419, 2018.

\bibitem{huang2019mask}
Zhida Huang, Zhuoyao Zhong, Lei Sun, and Qiang Huo.
\newblock Mask r-cnn with pyramid attention network for scene text detection.
\newblock In {\em 2019 IEEE Winter Conference on Applications of Computer
  Vision (WACV)}, pages 764--772. IEEE, 2019.

\bibitem{jiang2017r2cnn}
Yingying Jiang, Xiangyu Zhu, Xiaobing Wang, Shuli Yang, Wei Li, Hua Wang, Pei
  Fu, and Zhenbo Luo.
\newblock R2cnn: rotational region cnn for orientation robust scene text
  detection.
\newblock {\em arXiv preprint arXiv:1706.09579}, 2017.

\bibitem{karatzas2015icdar}
Dimosthenis Karatzas, Lluis Gomezbigorda, Anguelos Nicolaou, Suman~K Ghosh,
  Andrew~D Bagdanov, Masakazu Iwamura, Jiri Matas, Lukas Neumann, Vijay
  Chandrasekhar, Shijian Lu, et~al.
\newblock Icdar 2015 competition on robust reading.
\newblock pages 1156--1160, 2015.

\bibitem{kingma2014adam}
Diederik~P. Kingma and Jimmy Ba.
\newblock Adam: A method for stochastic optimization, 2014.

\bibitem{li2018shape}
Xiang Li, Wenhai Wang, Wenbo Hou, Ruo-Ze Liu, Tong Lu, and Jian Yang.
\newblock Shape robust text detection with progressive scale expansion network.
\newblock {\em arXiv preprint arXiv:1806.02559}, 2018.

\bibitem{li2018pixel}
Yuan Li, Yuanjie Yu, Zefeng Li, Yangkun Lin, Meifang Xu, Jiwei Li, and Xi Zhou.
\newblock Pixel-anchor: A fast oriented scene text detector with combined
  networks.
\newblock {\em arXiv preprint arXiv:1811.07432}, 2018.

\bibitem{liao2017textboxes}
Minghui Liao, Baoguang Shi, Xiang Bai, Xinggang Wang, and Wenyu Liu.
\newblock Textboxes: A fast text detector with a single deep neural network.
\newblock In {\em Thirty-First AAAI Conference on Artificial Intelligence},
  2017.

\bibitem{liao2018rotation}
Minghui Liao, Zhen Zhu, Baoguang Shi, Gui-song Xia, and Xiang Bai.
\newblock Rotation-sensitive regression for oriented scene text detection.
\newblock In {\em Proceedings of the IEEE Conference on Computer Vision and
  Pattern Recognition}, pages 5909--5918, 2018.

\bibitem{lin2017feature}
Tsung-Yi Lin, Piotr Doll{\'a}r, Ross Girshick, Kaiming He, Bharath Hariharan,
  and Serge Belongie.
\newblock Feature pyramid networks for object detection.
\newblock In {\em Proceedings of the IEEE conference on computer vision and
  pattern recognition}, pages 2117--2125, 2017.

\bibitem{liu2016ssd}
Wei Liu, Dragomir Anguelov, Dumitru Erhan, Christian Szegedy, Scott Reed,
  Cheng-Yang Fu, and Alexander~C Berg.
\newblock Ssd: Single shot multibox detector.
\newblock In {\em European conference on computer vision}, pages 21--37.
  Springer, 2016.

\bibitem{liu2018fots}
Xuebo Liu, Ding Liang, Shi Yan, Dagui Chen, Yu Qiao, and Junjie Yan.
\newblock Fots: Fast oriented text spotting with a unified network.
\newblock In {\em Proceedings of the IEEE conference on computer vision and
  pattern recognition}, pages 5676--5685, 2018.

\bibitem{long2015fully}
Jonathan Long, Evan Shelhamer, and Trevor Darrell.
\newblock Fully convolutional networks for semantic segmentation.
\newblock In {\em Proceedings of the IEEE conference on computer vision and
  pattern recognition}, pages 3431--3440, 2015.

\bibitem{long2018textsnake}
Shangbang Long, Jiaqiang Ruan, Wenjie Zhang, Xin He, Wenhao Wu, and Cong Yao.
\newblock Textsnake: A flexible representation for detecting text of arbitrary
  shapes.
\newblock In {\em Proceedings of the European conference on computer vision
  (ECCV)}, pages 20--36, 2018.

\bibitem{lyu2018multi}
Pengyuan Lyu, Cong Yao, Wenhao Wu, Shuicheng Yan, and Xiang Bai.
\newblock Multi-oriented scene text detection via corner localization and
  region segmentation.
\newblock In {\em Proceedings of the IEEE conference on computer vision and
  pattern recognition}, pages 7553--7563, 2018.

\bibitem{nayef2017icdar2017}
Nibal Nayef, Fei Yin, Imen Bizid, Hyunsoo Choi, Yuan Feng, Dimosthenis
  Karatzas, Zhenbo Luo, Umapada Pal, Christophe Rigaud, Joseph Chazalon, et~al.
\newblock Icdar2017 robust reading challenge on multi-lingual scene text
  detection and script identification - rrc-mlt.
\newblock pages 1454--1459, 2017.

\bibitem{ren2015faster}
Shaoqing Ren, Kaiming He, Ross Girshick, and Jian Sun.
\newblock Faster r-cnn: Towards real-time object detection with region proposal
  networks.
\newblock In {\em Advances in neural information processing systems}, pages
  91--99, 2015.

\bibitem{shi2017detecting}
Baoguang Shi, Xiang Bai, and Serge Belongie.
\newblock Detecting oriented text in natural images by linking segments.
\newblock In {\em Proceedings of the IEEE Conference on Computer Vision and
  Pattern Recognition}, pages 2550--2558, 2017.

\bibitem{tian2019learning}
Zhuotao Tian, Michelle Shu, Pengyuan Lyu, Ruiyu Li, Chao Zhou, Xiaoyong Shen,
  and Jiaya Jia.
\newblock Learning shape-aware embedding for scene text detection.
\newblock In {\em Proceedings of the IEEE Conference on Computer Vision and
  Pattern Recognition}, pages 4234--4243, 2019.

\bibitem{wang2018geometry}
Fangfang Wang, Liming Zhao, Xi Li, Xinchao Wang, and Dacheng Tao.
\newblock Geometry-aware scene text detection with instance transformation
  network.
\newblock In {\em Proceedings of the IEEE Conference on Computer Vision and
  Pattern Recognition}, pages 1381--1389, 2018.

\bibitem{wang2011end}
Kai Wang, Boris Babenko, and Serge Belongie.
\newblock End-to-end scene text recognition.
\newblock In {\em 2011 International Conference on Computer Vision}, pages
  1457--1464. IEEE, 2011.

\bibitem{xie2019scene}
Enze Xie, Yuhang Zang, Shuai Shao, Gang Yu, Cong Yao, and Guangyao Li.
\newblock Scene text detection with supervised pyramid context network.
\newblock In {\em Proceedings of the AAAI Conference on Artificial
  Intelligence}, volume~33, pages 9038--9045, 2019.

\bibitem{xue2018accurate}
Chuhui Xue, Shijian Lu, and Fangneng Zhan.
\newblock Accurate scene text detection through border semantics awareness and
  bootstrapping.
\newblock In {\em Proceedings of the European Conference on Computer Vision
  (ECCV)}, pages 355--372, 2018.

\bibitem{yang2018inceptext}
Qiangpeng Yang, Mengli Cheng, Wenmeng Zhou, Yan Chen, Minghui Qiu, Wei Lin, and
  Wei Chu.
\newblock Inceptext: A new inception-text module with deformable psroi pooling
  for multi-oriented scene text detection.
\newblock {\em arXiv preprint arXiv:1805.01167}, 2018.

\bibitem{yao2012detecting}
Cong Yao, Xiang Bai, Wenyu Liu, Yi Ma, and Zhuowen Tu.
\newblock Detecting texts of arbitrary orientations in natural images.
\newblock pages 1083--1090, 2012.

\bibitem{yao2016scene}
Cong Yao, Xiang Bai, Nong Sang, Xinyu Zhou, Shuchang Zhou, and Zhimin Cao.
\newblock Scene text detection via holistic, multi-channel prediction.
\newblock {\em arXiv preprint arXiv:1606.09002}, 2016.

\bibitem{zhang2019look}
Chengquan Zhang, Borong Liang, Zuming Huang, Mengyi En, Junyu Han, Errui Ding,
  and Xinghao Ding.
\newblock Look more than once: An accurate detector for text of arbitrary
  shapes.
\newblock In {\em Proceedings of the IEEE Conference on Computer Vision and
  Pattern Recognition}, pages 10552--10561, 2019.

\bibitem{zhang2016multi}
Zheng Zhang, Chengquan Zhang, Wei Shen, Cong Yao, Wenyu Liu, and Xiang Bai.
\newblock Multi-oriented text detection with fully convolutional networks.
\newblock In {\em Proceedings of the IEEE Conference on Computer Vision and
  Pattern Recognition}, pages 4159--4167, 2016.

\bibitem{zhong2019anchor}
Zhuoyao Zhong, Lei Sun, and Qiang Huo.
\newblock An anchor-free region proposal network for faster r-cnn-based text
  detection approaches.
\newblock {\em International Journal on Document Analysis and Recognition
  (IJDAR)}, 22(3):315--327, 2019.

\bibitem{EAST}
Xinyu Zhou, Cong Yao, He Wen, Yuzhi Wang, Shuchang Zhou, Weiran He, and Jiajun
  Liang.
\newblock East: an efficient and accurate scene text detector.
\newblock In {\em Proceedings of the IEEE conference on Computer Vision and
  Pattern Recognition}, pages 5551--5560, 2017.

\bibitem{zhu2018sliding}
Yixing Zhu and Jun Du.
\newblock Sliding line point regression for shape robust scene text detection.
\newblock In {\em 2018 24th International Conference on Pattern Recognition
  (ICPR)}, pages 3735--3740. IEEE, 2018.

\bibitem{zhu2018textmountain}
Yixing Zhu and Jun Du.
\newblock Textmountain: Accurate scene text detection via instance
  segmentation.
\newblock {\em arXiv preprint arXiv:1811.12786}, 2018.

\end{thebibliography}
}

\end{document}